\documentclass{article}
\pdfoutput=1

\usepackage{arxiv}

\usepackage[utf8]{inputenc} 
\usepackage[T1]{fontenc}    
\usepackage{hyperref}       
\usepackage{url}            
\usepackage{booktabs}       
\usepackage{amsfonts}       
\usepackage{nicefrac}       
\usepackage{microtype}      
\usepackage{lipsum}		
\usepackage{graphicx}
\usepackage{doi}
\usepackage[square, comma, sort&compress, numbers]{natbib}

\graphicspath{}

\title{Underwater Object Tracker: UOSTrack for Marine Organism Grasping of Underwater Vehicles}

\date{} 					

\author{YunFeng Li\\
	Harbin Engineering University\\
	\texttt{liyunfeng@hrbeu.edu.cn} \\
	\And
	Bo Wang\\
	Harbin Engineering University\\
	\texttt{cv\_heu@163.com} \\
 	\And
	Ye Li\\
	Harbin Engineering University\\
	\texttt{liye@hrbeu.edu.cn} \\
  	\And
	Wei Huo\\
	Harbin Engineering University\\
	\texttt{weihuo@hrbeu.edu.cn} \\
  	\And
	ZhuoYan Liu\\
	Harbin Engineering University\\
	\texttt{liuzhuoyan@hrbeu.edu.cn} \\
   	\And
	Yueming Li\\
	Harbin Engineering University\\
	\texttt{each-day@163.com} \\
   	\And
	Jian Cao\\
	Harbin Engineering University\\
	\texttt{caojian\_heu@163.com} \\
}

\renewcommand{\shorttitle}{\textit{arXiv} Template}

\hypersetup{hypertex=true,
            colorlinks=true,
            linkcolor=blue,
            anchorcolor=blue,
            citecolor=blue}

\begin{document}
\maketitle

\begin{abstract}
A visual single-object tracker is an indispensable component of underwater vehicles (UVs) in marine organism grasping tasks. Its accuracy and stability are imperative to guide the UVs to perform grasping behavior. Although single-object trackers show competitive performance in the challenge of underwater image degradation, there are still issues with sample imbalance and exclusion of similar objects that need to be addressed for application in marine organism grasping. This paper proposes Underwater OSTrack (UOSTrack), which consists of underwater image and open-air sequence hybrid training (UOHT), and motion-based post-processing (MBPP). The UOHT training paradigm is designed to train the sample-imbalanced underwater tracker so that the tracker is exposed to a great number of underwater domain training samples and learns the feature expressions. The MBPP paradigm is proposed to exclude similar objects. It uses the estimation box predicted with a Kalman filter and the candidate boxes in the response map to relocate the lost tracked object in the candidate area. UOSTrack achieves an average performance improvement of 4.41\% and 7.98\% maximum compared to state-of-the-art methods on various benchmarks, respectively. Field experiments have verified the accuracy and stability of our proposed UOSTrack for UVs in marine organism grasping tasks. More details can be found at https://github.com/LiYunfengLYF/UOSTrack. 
\end{abstract}


\section{Introduction}
Traditional manual fishing for marine organisms is associated with high risks and low efficiency. In contrast, a more promising trend is the use of underwater vehicles (UVs) to grasp marine organisms, and the UVs could promote efficiency, reduce costs and achieve long-term operation while ensuring operational safety. Visual single object tracking (SOT), typically used to locate and track targets in real-time, is an essential technology for UVs in marine organism grasping tasks, and its accuracy and stability are imperative in guiding the UV to complete the grasping behavior.

Underwater object tracking (UOT), as a branch of SOT, considers not only the challenges posed by SOT, but also the challenges posed by the underwater environment and the characteristics of marine targets. Most of the existing underwater trackers have been developed based on SOT trackers. Several approaches to improve feature extraction \cite{kcf_based_uot2} \cite{wu2023hybrid} and enhance tracker adaptability to appearance changes \cite{kcf_based_uot1} \cite{bacf_based_uot1} have been introduced to underwater trackers to address general challenges such as occlusion, appearance changes, etc. These efforts do not analyze the difference between UOT and SOT in detail, and although these improved methods effectively improve the performance of underwater trackers, this is done through more powerful features or better adaptability to model the similarity of underwater targets in the template and search area, without fully accounting for underwater image degradation or discriminative features of the underwater targets themselves. In addition, some underwater trackers \cite{uot100} \cite{ustark} are designed to address the underwater image degradation posed by the underwater environment by performing domain migration of images to reduce the gap between underwater and open-air images before using SOT trackers (also called open-air trackers) to locate targets more accurately. Such approaches are not about improvements to the tracker itself but about improving the quality of the underwater images themselves. In general, most previous work has focused on improving the tracker model and the input image, without fully considering the significant differences in the appearance model (including its discriminative features) and the behavior model between underwater and open-air targets. These two issues have not received much attention in previous studies, which limits the performance of existing underwater trackers.

The first problem is that, the training sets of existing underwater trackers primarily consist of large open-air datasets, which contain a limited number of underwater target samples. As a result, it is difficult for underwater trackers to effectively learn feature expressions specific to underwater targets. Therefore, underwater trackers face significant sample imbalance between the training and test sets, encompassing domain, class, and quantity imbalances. Since single-object trackers support all types of objects, sample imbalance is ignored not only in UOT but also in other specific object tracking approaches (such as UAV object tracking). There is currently no precedent for adjusting the imbalance between specific and general object samples. In addition, the lack of large-scale UOT training datasets and the high cost of tracking dataset collection and labelling make it difficult to adjust for the imbalance between underwater and open-air objects.

There are many available datasets for underwater object detection (UOD) and underwater object segmentation (UOS) and so on. Although the images in these datasets are not paired, their annotation densities are generally much higher than those of the UOT datasets (a detection image contains multiple targets, whereas a tracking sequence contains only one target). Therefore, it is useful to investigate the conversion of these non-sequential datasets into the tracker’s training set to adjust for the sampling imbalance of underwater trackers.

As for the second problem, targets such as fishes and dolphins often occur in groups, resulting in several objects with similar appearances being present in the search area. Despite the competitive performance of advanced open-air trackers and well-designed underwater trackers in addressing underwater image degradation problems, they fail to make sufficient distinction between similar objects because neural networks respond similarly to objects with comparable appearances.

Using MOSSE\cite{mosse} as a starting point, most trackers are developed by adding a Hanning window to the response map to suppress interference from similar distractors in the non-central regions of the search area. Then most trackers use well-designed networks to improve the target discrimination among similar objects. For example, KeepTrack \cite{keeptrack}provides an implicit motion modeling module that maintains tracking of all potential objects to discriminate between the tracked target and distractors. Although these methods play an important role in dealing with similar objects, they have limited performance gains, are computationally expensive, and cannot be quickly modified for a given tracker.

In some fields, the algorithms employed can naturally distinguish between similar instances. For example, in object detection, the detectors output all detection boxes. In multiple object tracking (MOT), the trackers match boxes in adjacent frames to record the motion information for each instance. The use of motion matching prevents each bounding box from jumping to similar objects. An important requirement is that the detector is sufficiently able to distinguish each instance. The tracker then matches the tracked targets based on the motion information in each frame. These two points can be fulfilled by SOT trackers. In most trackers, any area that resembles the template in the search area (called the candidate area) will score high on the response map. Trajectory matching could be introduced in SOT trackers. Therefore, it is a valuable exercise to exclude similar objects around a target based on motion matching.

This study includes the following research works:

1. The reason for the sample imbalance of OSTrack\cite{ostrack} is analysed for underwater object tracking. And underwater image and open-air sequence hybrid training (UOHT) is proposed. In particular, trainable sample pairs are constructed in the non-sequential datasets by using data augmentation methods to adjust the sample imbalance and learn the feature representation of underwater objects.

2. The reason for existing trackers tracking drift to similar objects is identified. And a motion-based post-processing (MBPP) paradigm is proposed. The MBPP uses trajectory prediction to relocate the tracked target hidden in candidate regions and corrects the corrects the target's tracking drift.

3. The effectiveness of UOHT and MBPP, and the generalization of MBPP, are verified by comparison experiments and ablation experiments. And UOSTrack sets a new state-of-the-art (SOTA) benchmark on the UOT100\cite{uot100} and UTB180\cite{utb180} datasets.

\section{Related Work}
\subsection{SOT tracker using detection data}

Previous research on tracker sample imbalance mainly focused on adjusting the positive and negative sample imbalances of models \cite{vital}\cite{lu2020deeploss}. There are few methods available for adjusting the imbalance between the number of specific target sample pairs and the number of general training samples. As far as we know, this is the first study to adjust for sample imbalance in specific (underwater) object tracking.

In addition, object detection datasets have been used to train trackers. DaSiamRPN \cite{dasiamrpn} introduced the ImageNet Detection\cite{imagenetdetection} and COCO\cite{COCO} datasets to construct positive and negative semantic samples based on different images, with a view to improving model discrimination of general targets. In contrast, our UOHT training paradigm requires a single underwater object detection image to construct sample pairs and adjust sample imbalance. UOD data are used to force the model to learn specific representations of target feature, thereby improving its discrimination of underwater objects.

ID-DSN\cite{IDDSN} was proposed for remote-sensing object tracking (RSOT). This approach uses only remote sensing detection datasets to construct sample pairs by image-clipping instance scaling to train the tracker. In RSOT, the target usually has no deformation. Therefore, it is feasible to train the tracker using only the detection dataset. In UOT, on the other hand, deformation is common and the model needs to learn the motion and deformation of the target. In contrast, in our UOHT, in addition to using underwater object detection datasets, common tracking datasets are also maintained, learning feature representations of underwater objects and modeling object motion and deformation.

\subsection{Interference suppression of similar objects in trackers}
A Hanning window was added to the response maps of the early trackers (MOSSE\cite{mosse}, KCF\cite{kcf}, DSST\cite{dsst}, SiamFC\cite{siamfc}) to suppress interference from similar targets at the search area boundary. This is the approach used by most trackers. However, the Hanning window cannot effectively suppress interference from the centre of the search area.

Improving model discrimination plays an important role in interference suppression. In the case of Siamese-based trackers, SiamRPN++ \cite{siamrpn++} uses deeper feature extraction networks to improve SiamRPN \cite{siamrpn}. In the case of Transformer-based trackers, Stark \cite{stark} employs a transformer for feature fusion and CornerHead for end-to-end bounding box prediction. MixFormer \cite{mixformer} uses a hybrid convolution and self-attention module to efficiently extract and fuse features. AiAtrack \cite{aiatrack} improves feature modeling by introducing self-attention in the self-attention module. OSTrack \cite{ostrack} employs VIT to jointly extract and fuse template and search area features to achieve SOTA performance.

Trackers that have a mechanism for self-updating templates can effectively maintain their adaptability to changes in the target’s appearance. Template background information discrimination and improved tracker discrimination are introduced through online updating in\cite{dimp}. KeepTrack\cite{keeptrack} tracks the positions of all potential targets by building a target candidate extraction network and association network that discriminates between the tracked target and distractors. TOMP\cite{tomp} provides improved feature modeling in the DCF paradigm through a transformer network.

Although these approaches effectively improve model discrimination, they require lengthy training and careful parameter tuning, which increases the cost of applying them to other trackers. In contrast, the proposed MBPP requires no training, fewer parameters (two in total), and can be easily applied to other trackers with candidate boxes.

\subsection{Underwater object tracking for UVs}

Some underwater trackers need to be deployed on UVs to perform target perception. Due to the limitations of energy and computing power, some traditional SOT trackers and Correlation Filtering (CF) trackers have been improved to achieve accurate tracking of underwater targets. Their improvement methods can be summarized as follow: improving the feature extraction of the model and utilizing template update mechanisms to enhance the appearance adaptability of the model. An adaptive appearance model and tracking strategy are used to improve KCF \cite{kcf_based_uot1}, and feature fusion and scale correction mechanisms are introduced to KCF\cite{kcf_based_uot2}. Besides, self-adaptive features, scale estimation and model updating mechanisms are used by BACF \cite{bacf_based_uot1}. In addition, TLD-based \cite{tld_based_uot1} trackers and particle-filter-based trackers \cite{pf_based_uot1l} \cite{pf_based_uot2} are used to track underwater targets. For some UVs with sufficient computing power, more powerful deep learning-based trackers are introduced to UOT. These trackers also focus on improving feature extraction from underwater targets. Such as, SiamRPN++ was improved using a reverse residual bottleneck block \cite{usiamrpn++}. SL-HENet proposed a global excitation model (HE module) to improve SiamRPN \cite{wu2023hybrid}. Unlike these underwater trackers, our UOHT effectively improves the learning of discriminative features for underwater targets from a data validity perspectiveand does not require additional time for inference.

In addition, more attention has been paid to improving the quality of underwater degraded images to improve tracker performance. As a GAN-based underwater image enhancement (UIE) method, CRN-UIE\cite{uot100} significantly improves the performance of various underwater trackers by improving the quality of underwater images. UStark\cite{ustark} employs image-adaptive enhancement to address various underwater image distortions. An underwater object tracking method with image enhancement and feature fusion (IEFF) is proposed to improve ECO\cite{wang2022underwater}. Although our tracker does not directly use underwater image enhancement methods, the introduction of a large number of underwater samples in UOHT makes the tracker itself better adapted to underwater domain images.

Some marine organisms have obvious social characteristics and exhibit aggregation behavior that differs from the motion behavior of open-air objects. Therefore, it is not reasonable for existing underwater trackers to simply follow the post-processing approach of the SOT tracker. Our MBPP paradigm fully considers the challenge of similar objects caused by the aggregation behavior, and uses the motion information to effectively exclude the interfering objects around the target.

\section{Experimental Underwater Vehicles Platform}
The UV used for the marine field experiments has an open frame structure with a 1.2 m three-joint manipulator. It is 1.2 m high, 0.6 m wide and 0.8 m long, and is capable of underwater navigation and autonomous operation. More details are shown in table \ref{tab:table1}. The visual subsystem of our UV for underwater object tracking consists of an underwater camera, an underwater LED light, and an embedded processing device (in our UV, an NVIDIA Jetson AGX Xavier module is used). The underwater camera is used to capture an image stream that is transferred to a processing device via a data switch. The UOT method is deployed in the embedded device, and the object locations in the tracking results are transferred to the control subsystem, which drives the UV to complete the grasping action with its manipulator. The manipulator of the UV uses image-based visual servoing (IBVS) scheme \cite{wang2018adaptive}. It uses a monocular camera for image processing and directly uses image errors to control the manipulator. By using the IBVS scheme, the manipulator no longer needs to estimate the relative position of the target.

\begin{figure}
	\centering
        \includegraphics[width=16cm]{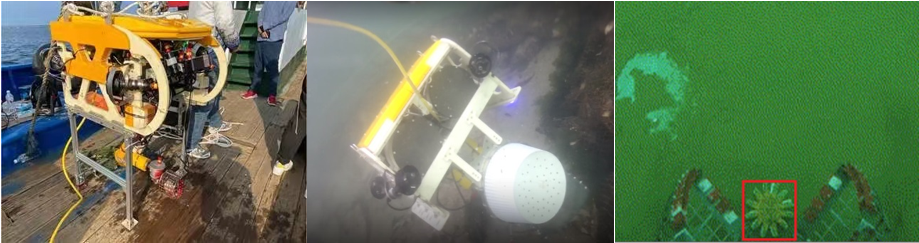}
	\caption{The UV for field experiments.}
	\label{fig:fig1}
\end{figure}

\begin{figure}
	\centering
        \includegraphics[width=16cm]{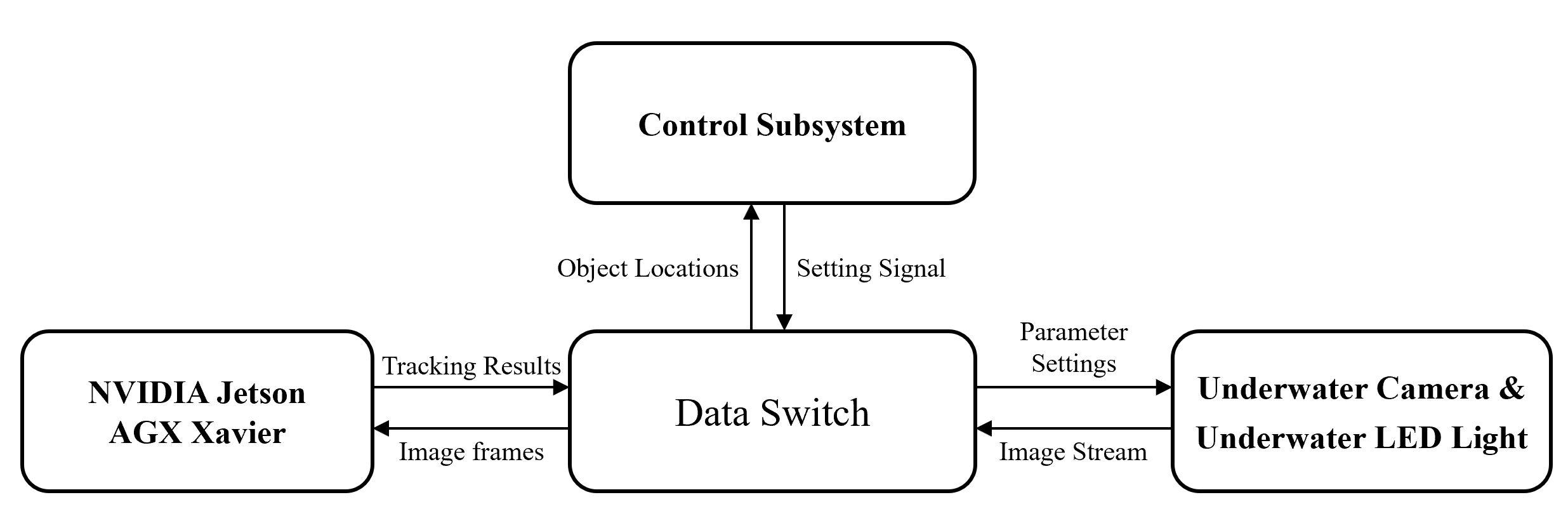}
	\caption{The structure of our UV’s visual subsystem.}
	\label{fig:fig2}
\end{figure}

\begin{table}[h]
\centering
\caption{Basic parameters of the UV}
\begin{tabular}{|c|c|}
\hline
Type                 & Parameter \\ \hline
length               & 0.8m      \\ \hline
width                & 0.6m      \\ \hline
height               & 1.2m      \\ \hline
Maximum diving depth & 100m      \\ \hline
UV weight            & 95.2kg    \\ \hline
Manipulator weight   & 15.4kg    \\ \hline
Total power          & 3500W     \\ \hline
Number of propellers & 6         \\ \hline
Maximum design speed & 1.5m/s    \\ \hline
Camera resolution    & 1920×1080 \\ \hline
Maximum illumination & 4000lux   \\ \hline
Field of vision      & 70º       \\ \hline
\end{tabular}
\label{tab:table1}
\end{table}

\section{Methods}
\label{sec:others}

OSTrack is a one-stream, transformer-based tracker that achieves SOTA performance in SOT. OSTrack consists of three parts: ViT \cite{vit}, which is used for joint feature extraction and relation modeling, Center-head \cite{ostrack} which is used for category and box prediction, and detection-based post-processing (DBPP) which is used for target localization. The training datasets for OSTrack included LaSOT\cite{lasot}, GOT10K \cite{got10k}, TrackingNet \cite{trackingnet}, and COCO \cite{COCO}.

\begin{figure}
	\centering
        \includegraphics[width=16cm]{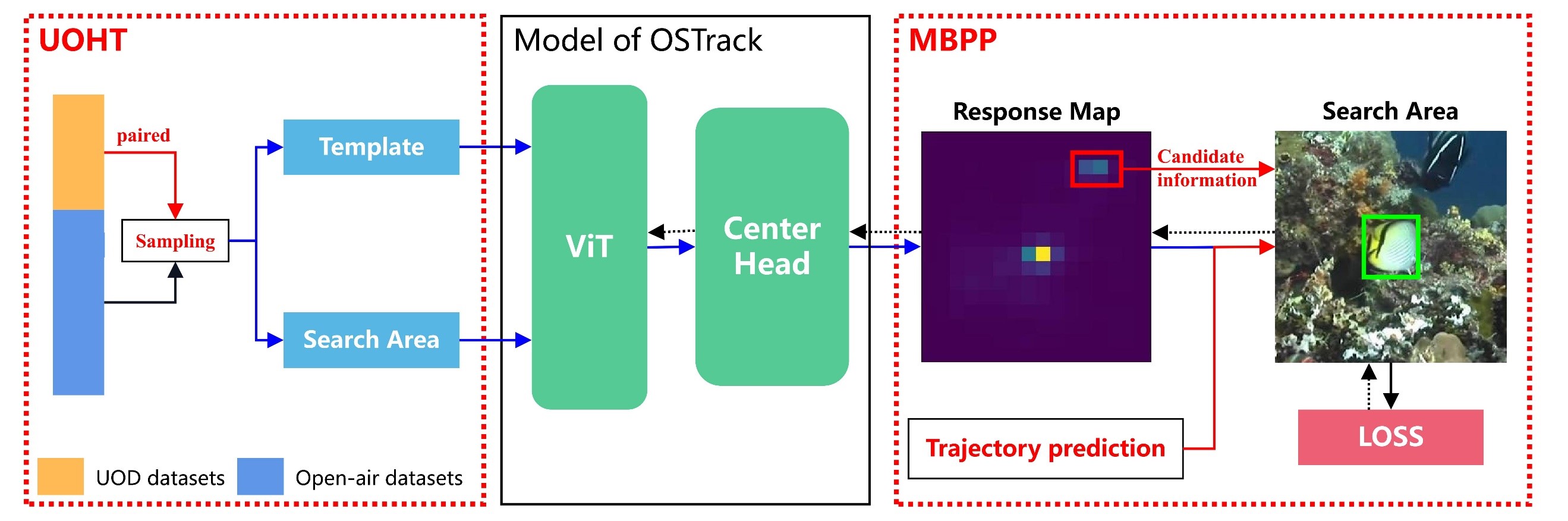}
	\caption{Overall framework of the proposed UOTrack system, and UOSTrack UOHT. During training, UOHT was used to construct high-quality underwater and open-air samples to train the model, thus ensuring that the tracker has less bias towards underwater objects. During tracking, MBPP was used to correct box drift when the model incorrectly locates similar objects.}
	\label{fig:fig3}
\end{figure}

To improve the performance of OSTrack\cite{ostrack}in UOT, the UOSTrack is proposed, which includes UOHT, and MBPP, as shown in Fig. \ref{fig:fig3}. UOSTrack follows the basic structure of OSTrack. The UOHT was designed to adjust the sample imbalance in OSTrack, and MBPP was designed to exclude similar objects around the target. It uses a monocular camera for image processing and directly uses image errors to control the manipulator. By using the IBVS scheme, the manipulator no longer needs to estimate the relative position of the target.

\subsection{Underwater images and open-air sequences hybrid training}

\subsubsection{Sample imbalance adjustment}

High quality training data are crucial for training a single object tracker in UOT. However, OSTrack for UOT has significant imbalances in terms of domain, class, and quantity of training datasets. There are 70 classes and 1400 sequences in LaSOT, but only 5 classes and 16 sequences are associated with underwater targets. GOT10K has 121 classes and 9335 sequences, but only 7 underwater classes and 284 underwater sequences. TrackingNet and COCO do not include underwater objects. Underwater trackers cannot learn high-quality generalized features with such a small number of underwater object classes and quantities. The large number of open-air samples leads to OSTrack to focus on learning the relation modeling and discrimination of open-air targets, and the losses of underwater objects are overwhelmed by the large number of open-air samples.

The sample imbalance, category imbalance, and quantity imbalance of the underwater tracker are unified into a sample imbalance and UOHT is used to train the tracker. UOHT converts unpaired UOD datasets into paired data (as discussed in section 4.1.2) and combines the paired data with open-air sequences to jointly train the tracker. The LaSOT, GOT10K, TrackingNet, and COCO datasets are reserved for learning relation-modeling. Two open UOD datasets (RUOD \cite{ruod} and FishExtend, which consists of two open datasets ) are added to balance the ratio of underwater samples to open-air samples. Another function of UOHT is to enable the model to learn the feature representations of underwater objects. The tracker not only relies on relational modeling to locate the target but also uses the features of underwater objects to identify and locate the target.

Adjusting for the sample imbalance, the ratio between the number of open-air and underwater samples for the UOSTrack training set is close to 2:1, while that for OSTrack is close to 99:1, as shown in Fig. \ref{fig:fig4}. 

Fig. \ref{fig:fig5} shows in detail how UOHT affects the target localization of the model. The heat map shows how much attention the model pays to the search area. The darker the heat map, the greater the contribution of that region to the box prediction. As shown in Fig. \ref{fig:fig5} (b-1), OSTrack focuses more on the boundaries of the shark, while UOSTrack focuses more on the entire shark in the background in Fig. \ref{fig:fig5} (c-1). The same situation is reflected in the comparison between Fig. \ref{fig:fig5} (e-1) and Fig. \ref{fig:fig5} (f-1), and between Fig. \ref{fig:fig5} (b-3) and Fig. \ref{fig:fig5} (c-3), and between Fig. \ref{fig:fig5} (e-3) and Fig. \ref{fig:fig5} (f-3). As shown in Fig. \ref{fig:fig5} (b-2), OSTrack focuses on any part of the search area that has similar features to dolphins when multiple dolphins congregate, while UOSTrack in Fig. \ref{fig:fig5} (c-2) can distinguish each individual dolphin and focus on the targets being tracked. Due to the sample imbalance, OSTrack generally relies on the similarity of features between each part of the search area and the template to locate a target. After learning the rich features of underwater targets, UOSTrack has a strong ability to locate underwater targets.

\begin{figure}[htbp]
    \begin{minipage}[t]{0.5\linewidth}
        \centering
        \includegraphics[width=7cm]{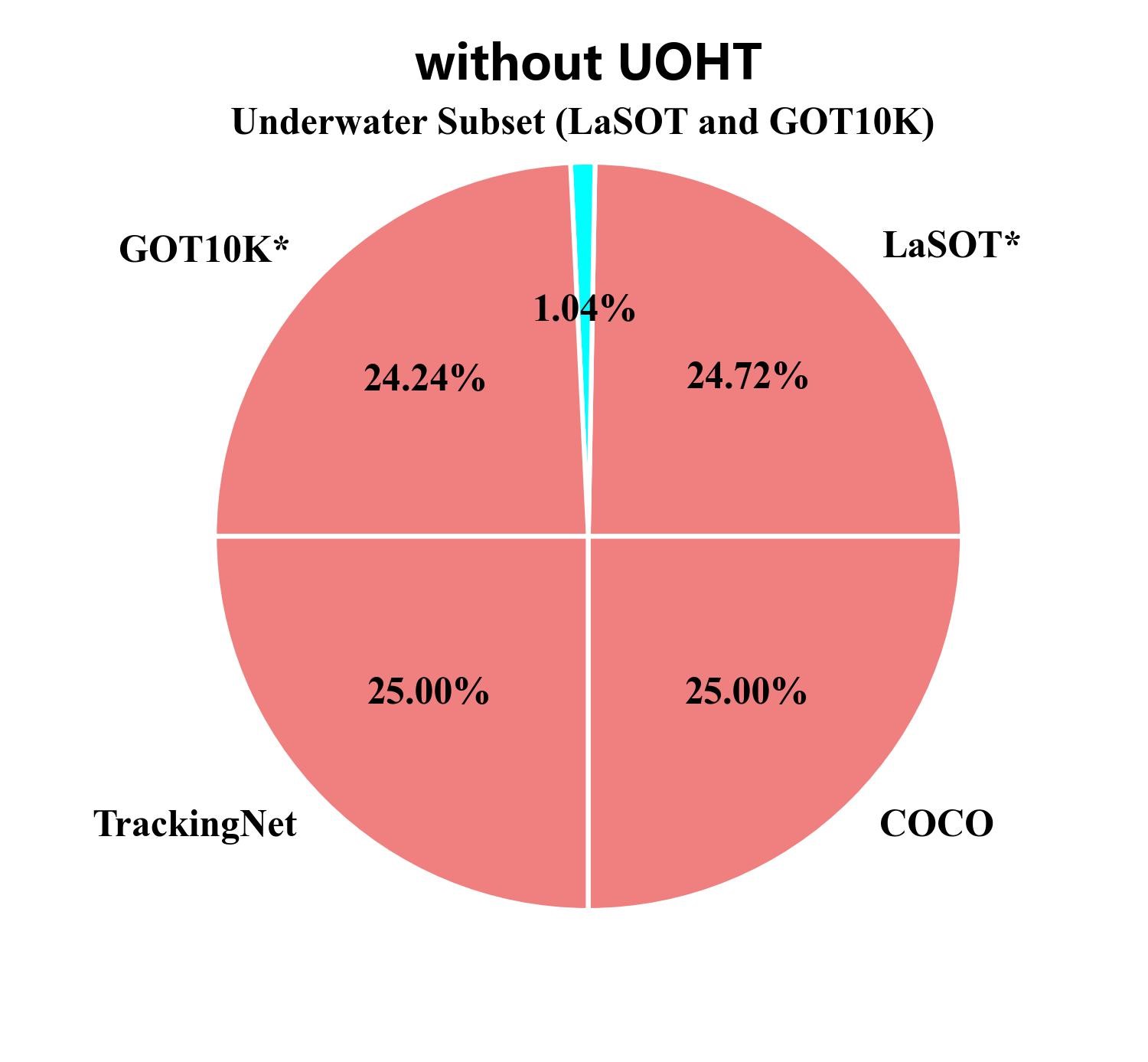}
    \end{minipage}%
    \begin{minipage}[t]{0.5\linewidth}
        \centering
        \includegraphics[width=7cm]{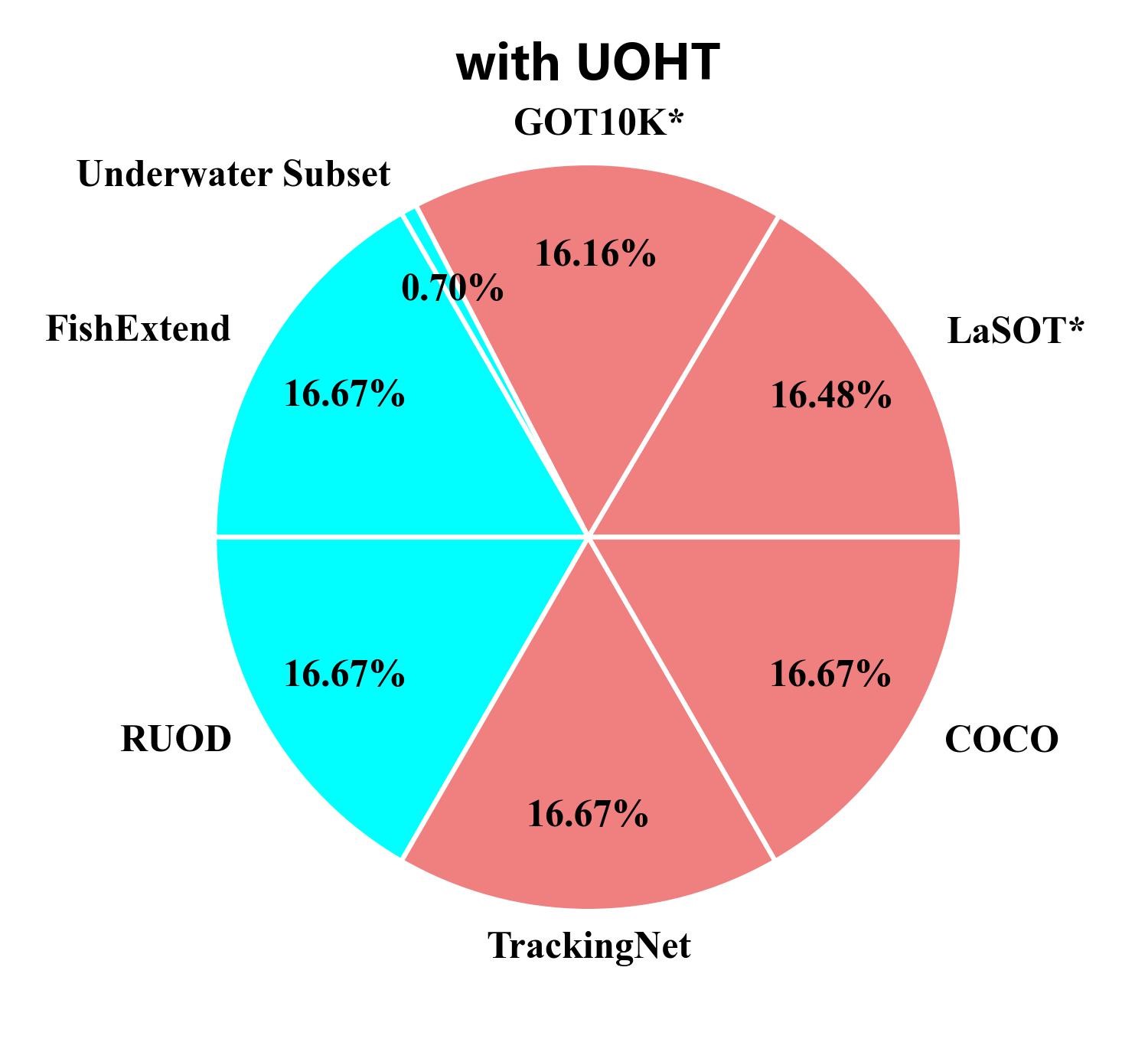}
    \end{minipage}
    \caption{Visualization of the proportion of each dataset sampled in an epoch. The cyan part indicates the underwater sample. The coral part indicates the open-air parts. (a) The proportion of samples in the OSTrack training set. (b) The proportion of samples in the UOSTrack training set.}
    \label{fig:fig4}
\end{figure}

\begin{figure}
	\centering
        \includegraphics[width=15cm]{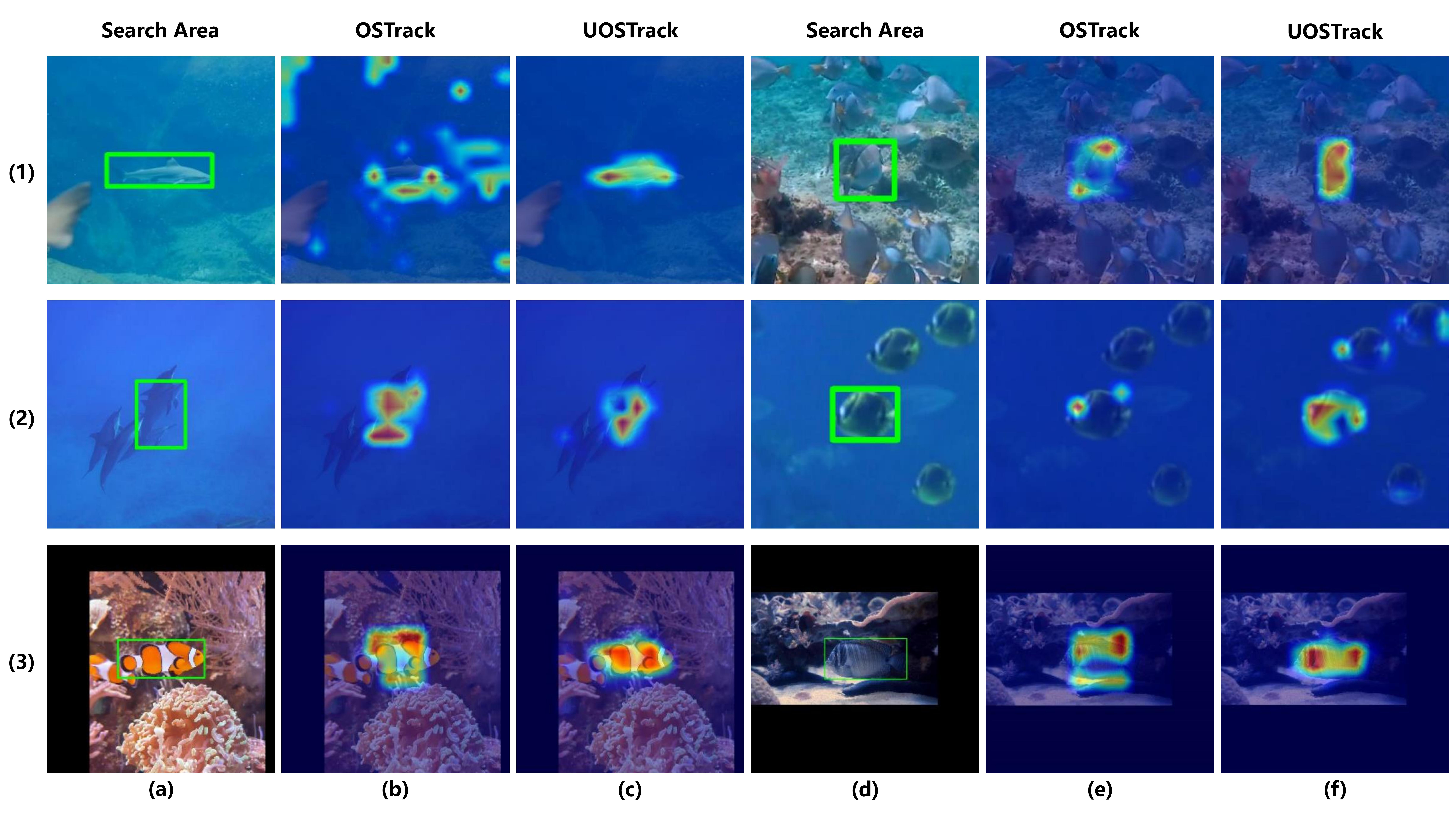}
	\caption{Visualization of discriminative regions of UOSTrack and OSTrack based on Grad-CAM. The green box represents the ground truth of the target in the current search area.}
	\label{fig:fig5}
\end{figure}

\subsubsection{Construct sample pairs of underwater images}

\begin{figure}
	\centering
        \includegraphics[width=15cm]{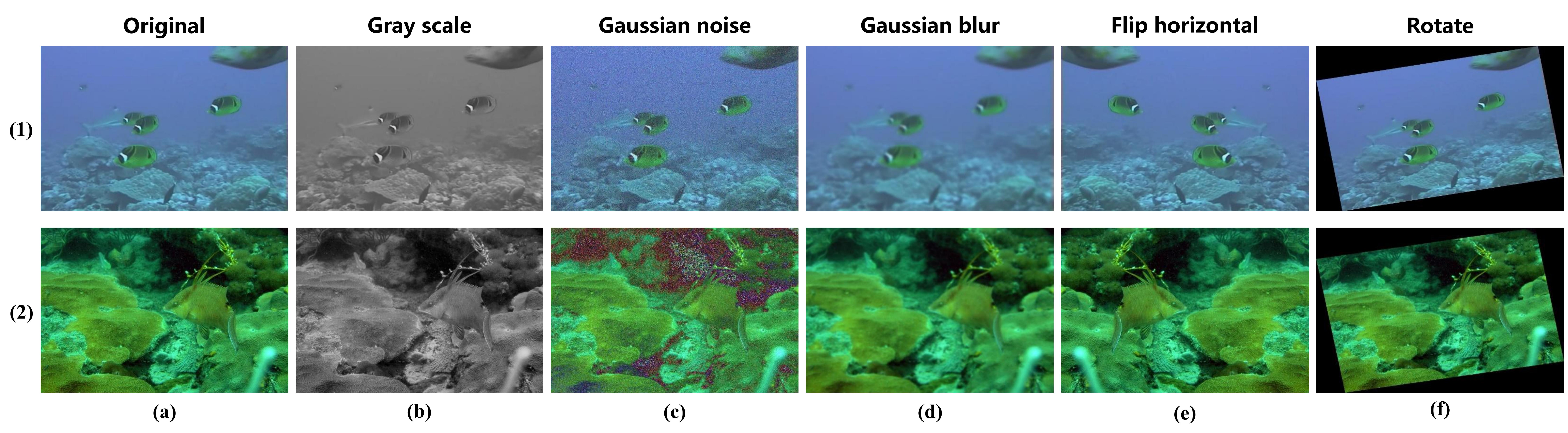}
	\caption{Visualization of discriminative regions of UOSTrack and OSTrack based on Grad-CAM. The green box represents the ground truth of the target in the current search area.}
	\label{fig:fig6}
\end{figure}

\begin{figure}
	\centering
        \includegraphics[width=10cm]{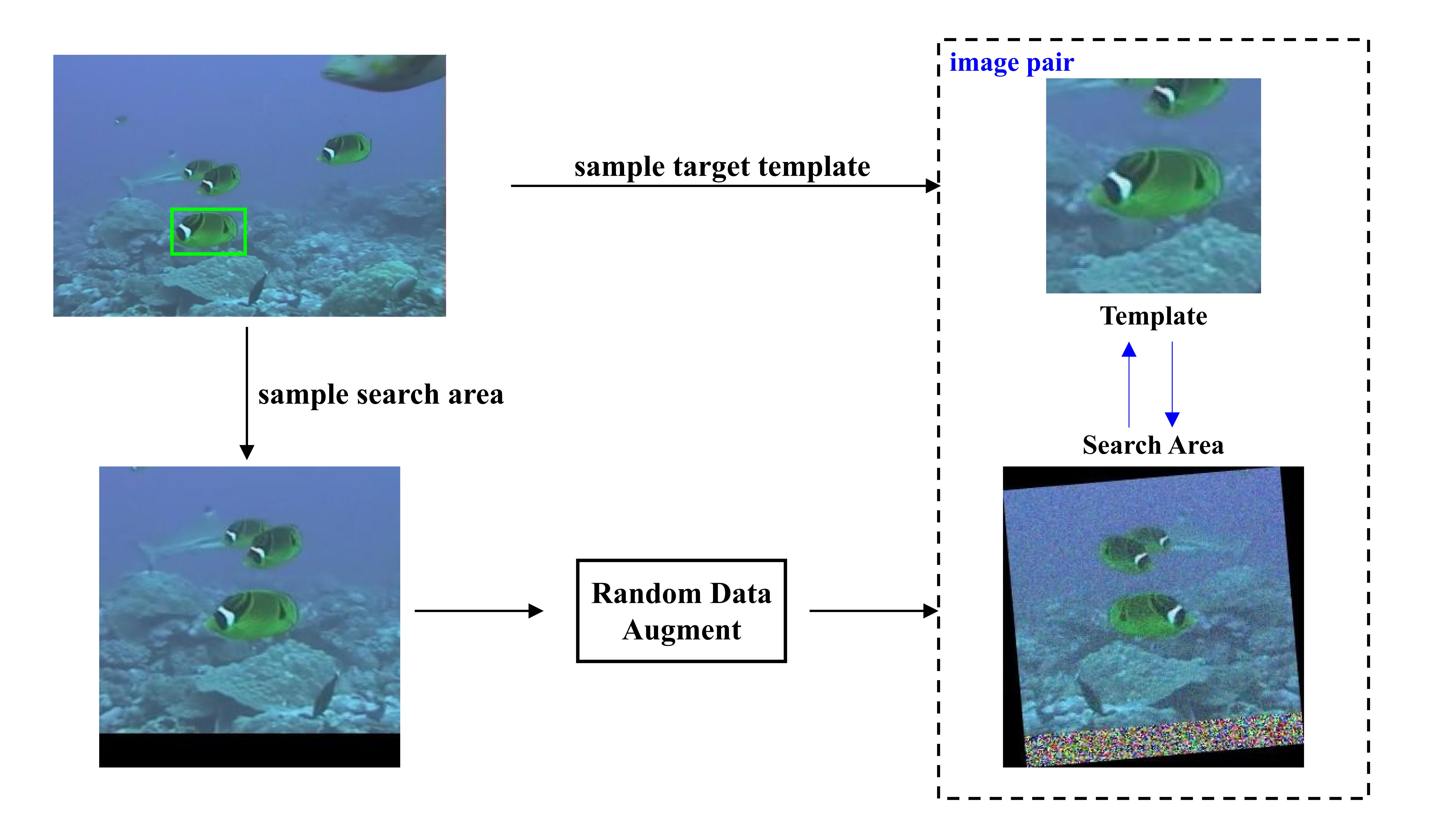}
	\caption{Visualization of discriminative regions of UOSTrack and OSTrack based on Grad-CAM. The green box represents the ground truth of the target in the current search area.}
	\label{fig:fig7}
\end{figure}

Image pairs are constructed for training since RUOD and FishExtend are unpaired detection datasets. The paired images are constructed by data augmentation. Following this paradigm, grayscale, horizontal flip, gaussian noise, gaussian blur, and rotation operations are used to construct the sample pairs. The effects of each data augmentation method are shown in Fig. \ref{fig:fig6}. The process of sample pair construction is shown in Fig. \ref{fig:fig7}. First, the target template and the search area are captured separately in the image, and then random data augmentation is used to augment the search area so that it can be used for training. In random data augmentation, each method of data augmentation independently determines whether to use it on its own probability threshold. Thresholds are greyscale at 0.1, horizontal flip at 0.15, gaussian noise at 0.05, gaussian blur at 0.05 and rotation at 0.05. The angle of rotation is between 0° and 10°.

\subsubsection{Hybrid training}
\begin{figure}
	\centering
        \includegraphics[width=10cm]{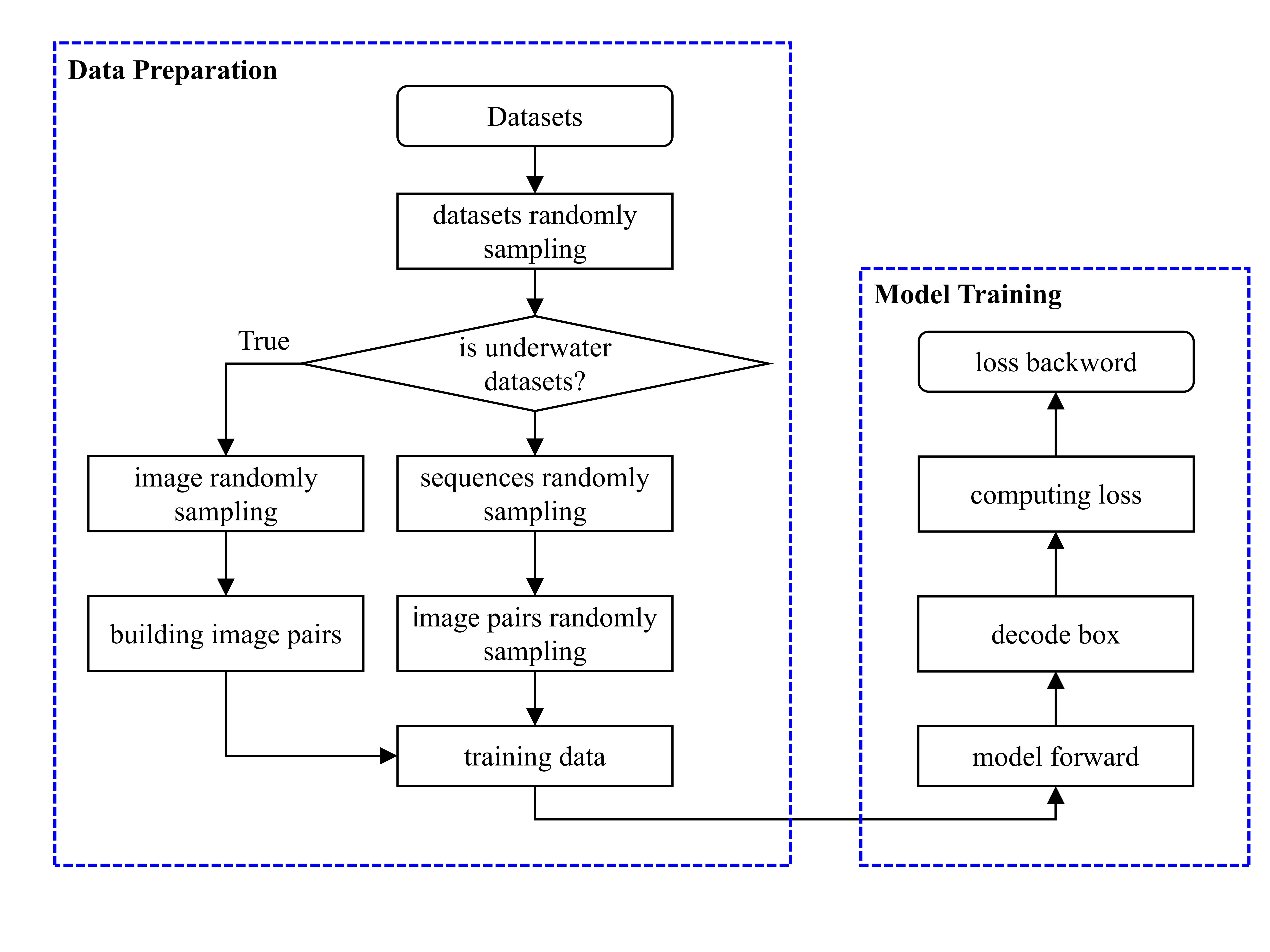}
	\caption{Visualization of discriminative regions of UOSTrack and OSTrack based on Grad-CAM. The green box represents the ground truth of the target in the current search area.}
	\label{fig:fig8}
\end{figure}

During the training process, an average sampling is first performed for 6 datasets. For the LaSOT, GOT10K, TrackingNet, and COCO datasets, the sampling mode remains consistent with the original OSTrack. For the RUOD and FishExtend datasets, an average sampling is performed for on the images and then a sample pair is constructed from a single image. After data preparation, the training data is fed into the model and the training is performed. The entire process is shown in Fig. \ref{fig:fig8}. 

The weight focal loss \cite{focalloss} is used for classification, the l1 loss and the generalized IoU loss \cite{giou} are used for box regression. The total loss function is:

\begin{equation}
    L_{total}=L_{cls}+\lambda_{iou}L_{iou}+\lambda{l_{1}}L_{1}
\end{equation}

where $\lambda_{iou}=2$ and $\lambda_{l_{1}}=5$ 

\subsection{Motion-based post-processing}
\subsubsection{Construction of candidate objects set}

The tracker uses the powerful appearance modeling ability of neural networks to distinguish targets. This means that any object that is similar to the target will have a similar response in the neural network. The set of these objects is collectively referred to as the candidate set.

OSTrack models a similarity metric between $f(z,x_{i})$ the template $z$ and the search area $x$ in the embedded  space $\varphi$

\begin{equation}
    f(z,x_{i})=\varphi(z,x)
\end{equation}

where $\varphi(z,x)$ denotes joint feature extraction and relation modeling of the template and the search area.

In frame $t$, the search area is divided into $s\times s$ patches, where $s$ is 16. Each patch then has a similarity score. Among the 256 patches, one is the tracked target $x_{t}$  and the others are all candidate patches. The candidate patches with the top-n similarity scores are extracted. At this time, the candidate set is denoted by $C_{t}:=\{\forall c_{i}\in C_{t}, f(z,c_{t})>h_{n}\}$, where $h_{n}$ is the $N_{th}$ largest similarity score. The candidate boxes set is $(B_{1},B_{2},...,B_{l})$, where l is the length of $C_{l}$. Then Non-maximum suppression (NMS) is then performed to further filter the redundant boxes for each candidate instance. 

\begin{equation}
    (c_{t}^{'},b)=\mathrm{NMS} (C_{t},B)
\end{equation}

The final candidate set is then collected as $C_{t}^{'}$, where $C_{t}^{,}\subset C_{t}$. The final candidate box  set is $(b_{1},b_{2},...,b_{n})$, where $n$ is the length of $C_{t}^{'}$, and $n<l$

The candidate boxes of the proposed tracker are visualized. Each object similar to the template has a high score on the response map, and each object is well covered by its bounding box, as shown in Fig.\ref{fig:fig4}.

\begin{figure}
	\centering
        \includegraphics[width=15cm]{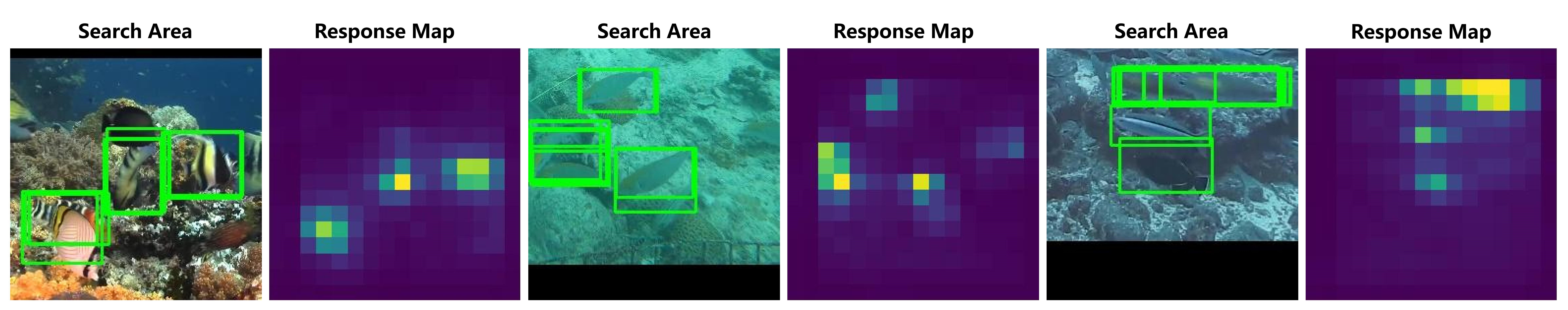}
	\caption{Visualisation of candidate information in response maps. Green boxes are the candidate boxes predicted by the network. Here, n is set to 10}
	\label{fig:fig4}
\end{figure}

\subsubsection{Target state prediction}

If the tracker suffers interference in the frame $k$, it may locate the interference target $\hat{x}_{k}$ due to its score being higher than the real target. In this time, the response of the real target will still exist in the response map, and $\hat{x}_{k},x_{k}\in C_{t}^{'}$. Since the tracker itself does not correct for incorrect positioning, $\hat{x}_{k}$ is output and $x_{k}$ is ignored. The goal of target state prediction is to predict the most likely position of $x_{k}$ in the frame $k$ so that the tracker can reposition the real target $x_{k}$ by using the real target’s predicted state.

The standard Kalman filter (KF) is introduced to predict the estimation box $b^{E}$  in each frame. The state and observation equations are as follows:

\begin{equation}
    X_{k}=AX_{k-1}+B_{\mu_{k-1}}+W_{k-1}
\end{equation}

\begin{equation}
    Z_{k}=HX_{k}+V_{k}
\end{equation}

where $X$ is the state vector of the target, $k$ represents time. $Z$ is the observation vector. $H$ is the observation matrix. $A$ and $B$  are the state transition and control matrices, respectively. $W$ and $V$ are the noise matrices that follow a Gaussian distribution with covariance matrices $Q$ and $R$, respectively.

Following the estimation model of SORT\cite{sort}, the target status is set as:

\begin{equation}
    X=[u,v,s,r,u^{'},v^{'},s^{'}]
\end{equation}

where $u$ and $v$ represent the horizontal and vertical positions of the target center respectively. $s$ represents the pixel area of the box. $r$ is aspect ratio of the box. $u^{'}$, $v^{'}$ and $s^{'}$ are the horizontal, vertical, area change speeds, respectively.

The time update functions used to predict the target state are as follows:

\begin{equation}
    \hat{X}^{-}_{k}=A\hat{X}_{k-1}+B\mu_{k-1}
\end{equation}

\begin{equation}
    {P}^{-}_{k}=AP_{k-1}A^{T}+Q
\end{equation}

where $\hat{X}^{-}_{k}$ is the priori estimate. ${P}^{-}_{k}$ is the error covariance. $\hat{X}_{k}$ and $P_{k-1}$ respectively denote the state vectors and covariance of errors at time $k-1$.

The estimation box $b^{E}$ is decoded by $\hat{X}^{-}_{k}=[\hat{u}^{-}_{k},\hat{v}^{-}_{k},\hat{s}^{-}_{k},\hat{r}^{-}_{k},\hat{u}^{‘-}_{k},\hat{v}^{’-}_{k},\hat{s}^{'-}_{k},]$ as:

\begin{equation}
    b^{E}=[\hat{u}^{-}_{k},\hat{v}^{-}_{k},\sqrt{\hat{s}^{-}_{k} / \hat{r}^{-}_{k}}, \sqrt{\hat{s}^{-}_{k}\times \hat{r}^{-}_{k}}]
\end{equation}

The measurement update functions which are used to update KF are as follows:
\begin{equation}
    K_{k}=P^{-}_{k}H^{T}(HP^{-}_{k}H^{T}+R)^{-1}
\end{equation}

\begin{equation}
    \hat{X}_{k}=\hat{X}^{-}_{k}+K_{k}(Z_{K}-H\hat{X}^{-}_{k})
\end{equation}

\begin{equation}
    {P}_{k}=(I-K_{k}H)P^{-}_{k}
\end{equation}
where $K_{k}$ is the gain matrix. $\hat{X}_{k}$ is the posteriori estimate. $P_{k}$ is the error covariance. $I$ is the identity matrix.

\subsubsection{Location score calculation}

The location score between $b^{E}$ and the candidate box $(b_{1},b_{2},...,b_{n})$is calculated. The maximum score indicates that that  candidate box is the  closest to the tracked target box.

\begin{equation}
score=f(z,c_{i})\times \mathrm{IoU} (box1, box2)
\end{equation}

where $f(z,c_{i})$ denotes the similarity score of $\mathrm{ith}$ candidate box, and $\mathrm{IOU}$ denotes the intersection over union (IoU) of the two boxes. .

\subsubsection{Matching process}

\begin{figure}
	\centering
        \includegraphics[width=15cm]{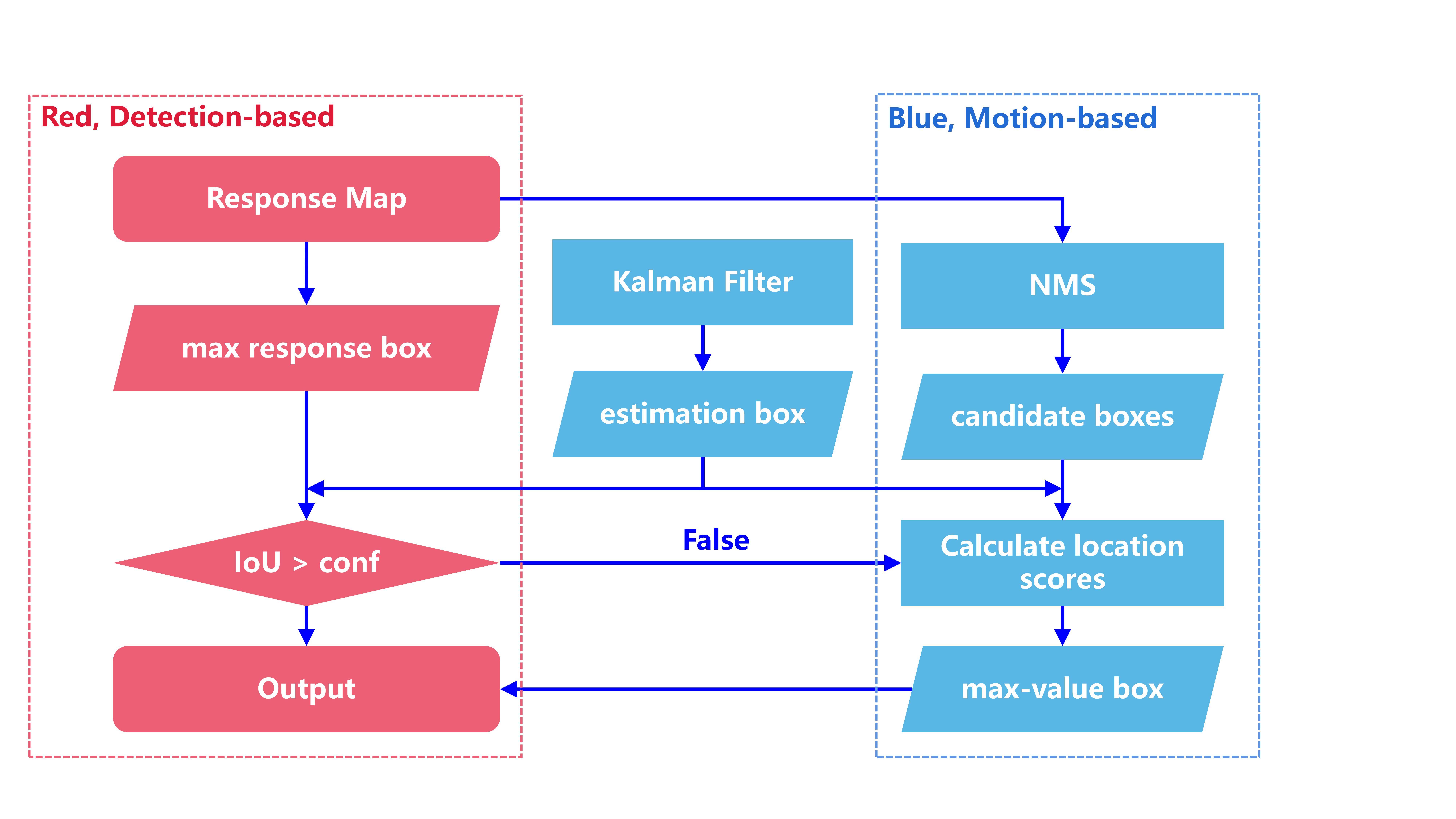}
	\caption{The matching process of MBPP.}
	\label{fig:fig10}
\end{figure}

The tracker initializes an additional KF in the first frame to record, update, and predict the target’s motion. As shown in Fig. \ref{fig:fig10}, during tracking, the tracker first follows detection-based post-processing and predicts the maximum response target $\tilde{x}$, and decodes its box. It then employs the KF to predict the estimation box $b^{E}$. The IoU is then calculated for max-response boxes $b_{ \tilde{x} }$ and $b^{E}$. If the IoU is less than $conf$, $\tilde{x}$ is $\hat{x}$, and motion-based match processing is used to locate the tracked target $x_{t}$ hidden in candidate set  $C_{t}^{'}$. Candidate boxes $(b_{1},b_{2},...,b_{n})$ are extracted from the response map and the location scores of $b_{i}$ and $b^{E}$ are calculated. The target with the highest score is then output as the tracked target $x_{t}$.

\section{Experiment}
\subsection{ Implementation details of simulation }

UOSTrack trains a total of 300 epochs on 2 NVIDIA RTX A6000. The batch-size is set to 32. The learning rate is set to 0.004. The optimizer is set to AdamW. At the 240th epoch, the learning rate decays to 0.0004.
The number of candidate boxes (n) is set to 40 for the trackers (OSTrack, Stark, and TransT) using the UOT100 dataset. A value of n = 30 is set for the UTB180 dataset. The conf value is set to 0.6 in the MBPP match processing. As discussed in Section 4.4, Stark’s CornerHead is replaced by CenterHead, the remaining parameters are transferred, and 50 epochs are trained according to the OSTrack training paradigm. In the simulation, the proposed method is first tested on a personal computer with an i9-12900KF CPU, a GX-3090Ti GPU, 16GB DDR4 memory, and PyTorch version 1.7.1 installed.

\subsection{Underwater Object Tracking Benchmark}

UOT100\cite{uot100} is a typical UOT benchmark dataset. Its sequences mainly reflect the color distortion of underwater images in different scenes. 28 subsets of UOT100 with similar objects are extracted as similar subsets of UOT (see Appendix A for details). The remaining 78 sequences are used simultaneously as complements to similar subsets.

UTB180\cite{utb180} is a high-quality UOT benchmark dataset. In addition to reflecting the color distortion of underwater images, the different subsets in UTB180 are divided according to nine challenges: scale variation, out-of-view, partial occlusion, full occlusion, deformation, low resolution, fast motion, motion blur, and similar objects.

Three metrics, success (AUC), precision (P), and norm-precision (P-NORM) are utilized in the evaluation of the tracker, which adheres to the one-pass evaluation (OPE) protocol that is most frequently used in SOT.

\subsection{Comparison between UOSTrack and SOTA trackers}

We evaluated 14 Open-air visual SOTA trackers (NeighborTrack\cite{ostrackN}, OSTrack\cite{ostrack}, MixFormer\cite{mixformer}, AiATrack\cite{aiatrack}, ToMP\cite{tomp}, KeepTrack\cite{keeptrack}, Stark\cite{stark}, TransT\cite{transt}, TrDimp\cite{trdimp}, SiamBAN-ACM\cite{siambanacm}, Dimp\cite{dimp}, SiamBAN\cite{siamban}, ATOM\cite{atom}, SiamCAR\cite{siamcar}) and UOSTrack on the UOT100 and UTB180 datasets. The published results for five classic open-air trackers (TLD\cite{tld}, KCF\cite{kcf}, ECO\cite{eco}, BACF\cite{bacf}, SiamRPN++\cite{siamrpn++}) commonly used in UOT over the past year are also compared for the UOT100 dataset. These trackers have been improved in UOT and the corresponding results may be found in the paper on UOT100\cite{uot100}. The results for underwater trackers published using the UOT100 and UTB180 datasets (only UStark) are also compared. These results are presented in Table \ref{tab:table2} and Table \ref{tab:table3}. 

\subsubsection{Results on the UOT100 dataset benchmark}
UOSTrack offers state-of-the-art performance, surpassing UStark by 0.44\% in terms of AUC and NeighborTrack by 0.59\% and 0.39\%. Additionally, it outperformed OSTrack in terms of AUC, P, and P-NORM by 1.36\%, 1.04\%, and 1.07\%, respectively. In a similar subset, UOSTrack outperformed OSTrack in terms of AUC, P, and P-NORM, respectively, by 3.47\%, 3.89\%, and 4.3\%.

\subsubsection{Results on the UTB180 dataset benchmark}
UOSTrack represents state-of-the-art performance and outperforms NeighborTrack in terms of AUC, P, and P-NORM, respectively, by 2.18\%, 2.15\%, and 2.86\%. Additionally, UOSTrack outperformed OSTrack in terms of AUC, P, and P-NORM by 3.08\%, 3.73\%, and 4.41\%, respectively. In a similar subset, UOSTrack outperformed OSTrack in terms of AUC, P, and P-NORM, respectively, by 6.27\%, 5.82\%, and 7.54\%.

\subsubsection{Performances of other trackers in UOT}

The performance of the improved SiamRPN++ is 20\% worse than the open-air OSTrack. This means that the widely used underwater tracker is unable to cope with the challenges of underwater tracking.
Another key phenomenon is that the P-NORM for modern OSTrack and Mixformer is close to 90\% for the components of similar subsets of UOT100 and UTB180. These trackers provide sufficient discrimination for general underwater tracking. However, for a similar subset, the accuracy and success rate of the tracker remain low, further proving that the challenge of similar objects plays a key role in UOT testing.

In addition, the P-NORM of these trackers is much higher than the AUC. This is because underwater image sequences face greater challenges from various boundary conditions (such as image degradation, occlusion, similar targets, etc.). They significantly lower the AUC of the tracker. Therefore, when applying the tracker to the UV, various boundary conditions need to be considered in order to achieve performance improvements, such as the low response score due to occlusion or out-of-view, the high confidence update scheme for updating the template or target state (such as MBPP), and its self-correction mechanism. Effective use of external information to improve tracking performance is also an important issue. In general, the design of a multi-strategy UOT method is valuable for promoting the task efficiency of UVs.

\begin{table}[h]\scriptsize
	\caption{Comparisons of UOSTrack with SOTA performance trackers on the UOT100. The two best results are shown in red and blue. (*denotes a tracker has improved in UOT))}
	\centering
	\begin{tabular}{cccccccccccc}
  	\cmidrule(r){1-12}	
        \multicolumn{3}{c}{\ } & \multicolumn{3}{c}{UOT100} &\multicolumn{3}{c}{UOT100’s Similary Subset}&\multicolumn{3}{c}{Complements of Similar Subsets}\\
 	\cmidrule(r){1-12}
 	\cmidrule(r){1-12}
        Tracker & Underwater tracker & Source & AUC & P & P-Norm & AUC & P & P-Norm & AUC & P & P-Norm \\
 	\cmidrule(r){1-12}
        TLD* &used & TPAMI2012 & 17.00 & 10.50 & - & -  & -  & -  & -  & -  & -  \\
        KCF* &used & TPAMI2014 & 30.10 & 18.90 & - & -  & -  & -  & -  & -  & -   \\
        ECO* &used & CVPR2017 & 37.90 & 27.80 & - & -  & -  & -  & -  & -  & -   \\
        BACF* &used & CVPR2017 & 38.70 & 38.70 & - & -  & -  & -  & -  & -  & -   \\
        SiamRPN++* &used & CVPR2019 & 54.00 & 46.20 & - & -  & -  & -  & -  & -  & -   \\
        \cmidrule(r){1-12}
        ATOM &\ & CVPR2019 & 54.79 & 44.24 & 68.76 & 48.63 & 41.42 & 59.84 & 57.48 & 46.03 & 73.53 \\
        Dimp &\ & ICCV2019 & 59.82 & 48.90 & 75.39 & 56.59 & 45.62 & 68.80 & 61.42 & 50.91 & 78.38 \\
        SiamCAR &\ & CVPR2020 & 53.55 & 45.96 & 69.40 & 44.14 & 40.88 & 54.78 & 57.32 & 48.44 & 75.06 \\
        SiamBAN &\ & CVPR2020 & 56.72 & 50.71 & 73.41 & 53.76 & 49.56 & 67.64 & 57.92 & 51.89 & 76.11 \\
        SiamBAN-ACM &\ & CVPR2021 & 61.43 & 51.89 & 75.01 & 55.84 & 48.07 & 65.70 & 63.73 & 54.05 & 78.82 \\
        TrDimp &\ & CVPR2021 & 61.19 & 51.04 & 77.47 & 54.90 & 44.42 & 66.32 & 63.66 & 54.17 & 81.92 \\
        TransT &\ & CVPR2021 & 63.75 & 56.27 & 79.85 & 57.12 & 50.02 & 67.74 & 66.23 & 59.03 & 84.28 \\
        Stark &\ & ICCV2021 & 66.33 & 58.12 & 82.66 & 57.50 & 49.82 & 68.03 & 69.63 & 61.68 & 88.04 \\
        UStark &\checkmark & JEI2022 & \color{blue}67.80 & 59.80 & - & - & - & - & - &- & - \\
        KeepTrack &\ & ICCV2021 & 60.04 & 51.20 & 78.05 & 54.48 & 46.29 & 66.84 & 63.02 & 53.66 & 82.37 \\
        ToMP &\ & CVPR2022 & 66.84 & 58.55 & 82.82 & \color{blue}61.94 & 53.69 & \color{blue}73.80 & 68.87 & 60.92 & 86.47 \\
        AiaTrack &\ & ECCV2022 & 65.31 & 57.57 & 83.06 & 59.67 & 52.24 & 72.43 & 67.48 & 60.04 & 87.02 \\
        MixFormer &\ & CVPR2022 & 66.20 & 59.81 & 83.50 & 58.97 & 50.89 & 69.78 & 68.94 & 63.55 & 88.63 \\
        OSTrack&\ & ECCV2022 & 66.88 & 62.11 & 84.55 & 58.81 & 54.33 & 72.08 & 69.88 & 65.49 & \color{blue}89.23 \\
        NeighborTrack &\ & ArXiv2022 & 67.32 & \color{blue}62.56 & \color{blue}85.23 & 59.82 & \color{blue}55.25 & 73.63 & \color{blue}70.15 & \color{red}65.79 & \color{red}89.66 \\        
  	\cmidrule(r){1-12}
        UOSTrack &\checkmark & \ & \color{red}68.24 & \color{red}63.15 & \color{red}85.62 & \color{red}62.28 & \color{red}58.22 & \color{red}76.38 & \color{red}70.50 & \color{blue}65.53 & 89.09 \\
  	\cmidrule(r){1-12}
	\end{tabular}
	\label{tab:table2}
\end{table}

\begin{table}[h]\scriptsize
	\caption{Comparisons of UOSTrack with SOTA performance trackers on the UTB180. The two best results are shown in red and blue.}
	\centering
	\begin{tabular}{cccccccccccc}
  	\cmidrule(r){1-12}	
        \multicolumn{3}{c}{\ } & \multicolumn{3}{c}{UTB180} &\multicolumn{3}{c}{UTB180’s Similary Subset}&\multicolumn{3}{c}{Complements of Similar Subsets}\\
 	\cmidrule(r){1-12}
 	\cmidrule(r){1-12}
        Tracker & Underwater tracker & Source & AUC & P & P-Norm & AUC & P & P-Norm & AUC & P & P-Norm \\
 	\cmidrule(r){1-12}
        ATOM &\ & CVPR2019 & 47.49 & 35.23 & 55.16 & 42.02 & 33.66 & 48.25 & 57.40 & 38.09 & 67.32 \\
        Dimp &\ & ICCV2019 & 50.52 & 37.51 & 58.49 & 43.41 & 34.48 & 49.33 & 63.39 & 43.01 & 75.08  \\
        SiamCAR &\ & CVPR2020 & 49.80 & 40.84 & 60.07 & 46.62 & 41.71 & 56.27 & 55.57 & 39.25 & 66.94 \\
        SiamBAN &\ & CVPR2020 & 56.95 & 46.95 & 68.22 & 54.29 & 47.11 & 64.53 & 61.77 & 46.64 & 74.90  \\
        SiamBAN-ACM &\ & CVPR2021  & 56.97 & 46.97 & 66.74  & 52.61 & 45.50 & 60.69 & 54.88 & 49.63 & 77.61 \\
        TrDimp &\ & CVPR2021 & 59.00 & 47.52 & 68.65 & 52.03 & 43.49 & 59.66 & 71.62 & 54.82 & 84.93 \\
        TransT &\ & CVPR2021 & 57.52 & 50.30 & 66.13 & 49.88 & 45.39 & 56.94 & 71.37 & 59.20 & 82.79 \\
        Stark &\ & ICCV2021 & 55.86 & 4845 & 64.63 & 45.75 & 40.85 & 52.13 & 74.18 & 62.21 & 87.27\\
        KeepTrack &\ & ICCV2021 & 54.90 & 43.49 & 64.06 & 45.71 & 37.78 & 52.52 & 71.55 & 53.82 & 84.99 \\
        ToMP &\ & CVPR2022 & 61.14 & 53.40 & 70.82 & 54.39 & 48.72 & 62.45 & 73.36 & 61.87 & 85.99 \\
        AiaTrack &\ & ECCV2022 & 62.01 & 52.94 & 72.11 & 55.14 & 48.13 & 63.39 & 74.47 & 61.67 & 87.91\\
        MixFormer &\ & CVPR2022 & 57.44 & 50.60 & 65.69 & 46.37 & 41.01 & 51.96 & \color{blue}77.49 & \color{red}68.00 & \color{blue}90.58  \\
        OSTrack&\ & ECCV2022 & 63.03 & 56.52 & 72.61 & 54.78 & 50.29 & 62.50 & \color{red}77.97 & \color{blue}67.83 & \color{red}90.92\\ 
        NeighborTrack &\ & ArXiv2022 & \color{blue}64.53 & \color{blue}58.10 & \color{blue}74.16 & \color{blue}58.21 & \color{blue}53.85 & \color{blue}66.29 &75.99 & 65.79 &88.42 \\  
  	\cmidrule(r){1-12}
        UOSTrack &\checkmark & \ & \color{red}66.71 & \color{red}60.25 & \color{red}77.02 & \color{red}61.05 & \color{red}56.14 & \color{red}70.04 & 76.96 &67.69 & 89.68 \\
  	\cmidrule(r){1-12}
	\end{tabular}
	\label{tab:table3}
\end{table}

\subsubsection{Lack of instance-level discrimination for close similar targets}
\begin{figure}
	\centering
        \includegraphics[width=15cm]{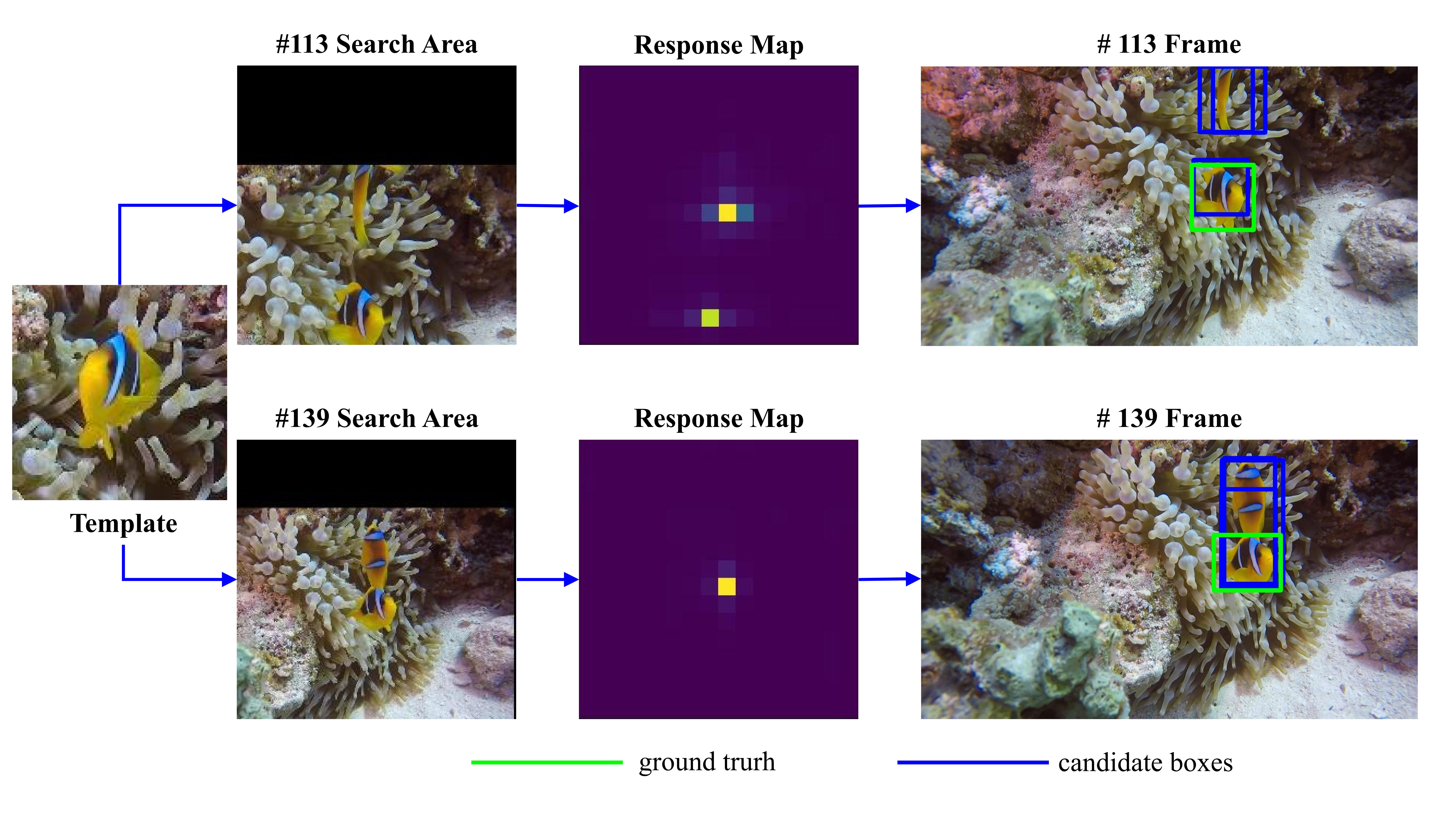}
	\caption{Illustrations of the lack of instance-level discrimination for close similar targets using the proposed tracker.}
	\label{fig:fig11}
\end{figure}

The proposed tracker is unable to discriminate between highly similar closed targets at the instance level. In frame 113 of Fig. \ref{fig:fig11}, there is a distance between two targets. These two targets are identified in the response map. However, the bounding box of the maximum response and its surrounding candidate boxes cover the two targets. The response map and the bounding boxes in frame 139 treat the frame as if it contains one large target when the targets are close together.

This lack of discrimination comes from the fact that the tracker obtains the final bounding box by matching features in the template with those in the search area. Each successfully matched feature is considered part of the output box. These are combined to form a large bounding box containing multiple similar objects. This bounding box effect degrades the performance of the tracker and can lead to tracking drift. It also contaminates the KF used in our method and affects the effectiveness of the MBPP.

\subsection{Effectiveness and Generalization of MDPP}
\begin{figure}
	\centering
        \includegraphics[width=14cm]{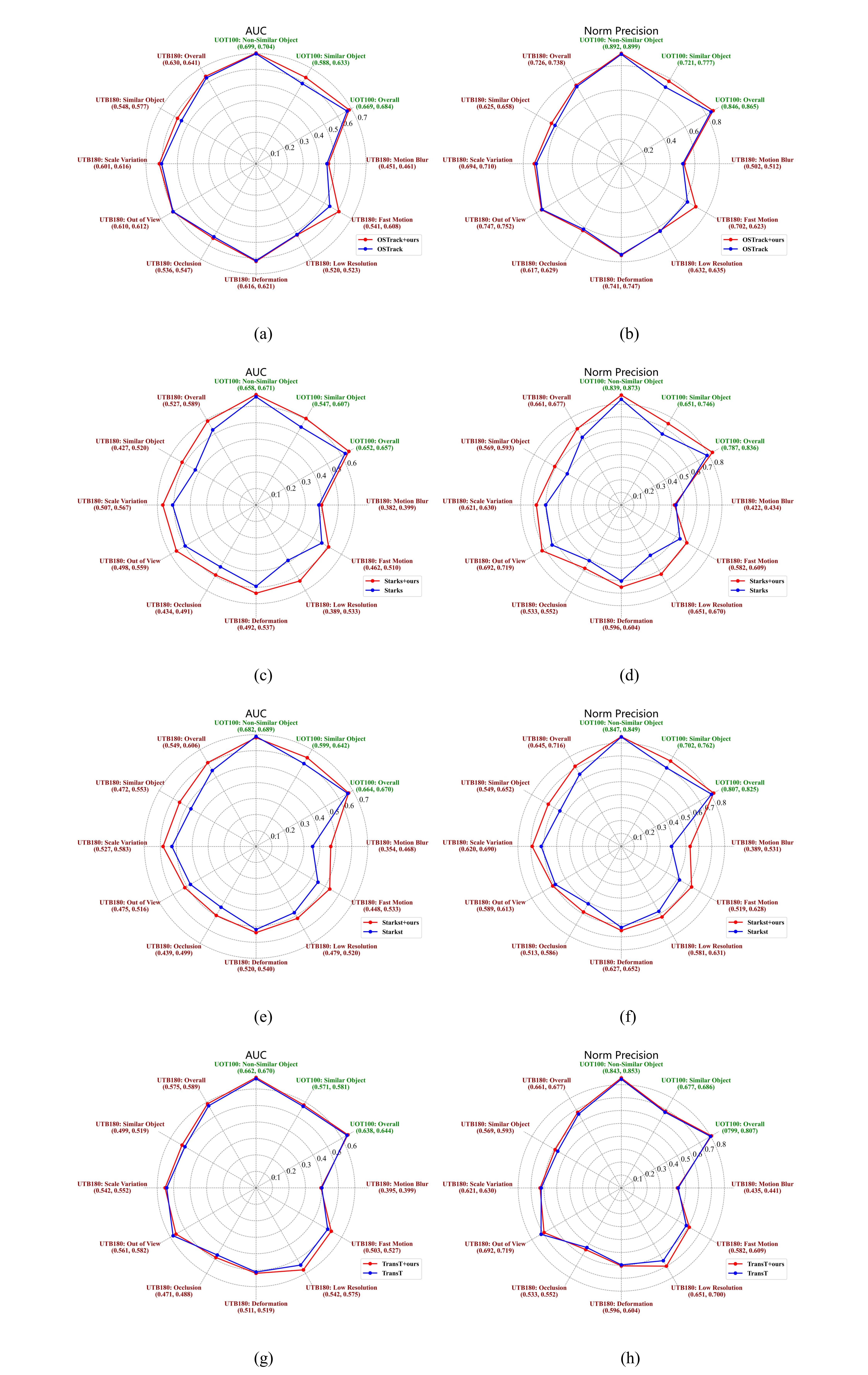}
	\caption{Comparison of AUC and P-NORM of trackers on UOT100 and UTB180 datasets for different visual attributes. Format for each attribute: \emph{name (min-value, max-value)}. (a) and (b) The AUC and Norm-Precision radar map of OSTrack and OSTrack+MBPP. (c) and (d) The AUC and Norm-Precision radar map of Starks and Starks+MBPP. (e) and (f) The AUC and Norm-Precision radar map of Starkst and Starkst +MBPP. (g) and (h) The AUC and Norm-Precision radar map of TransT and TransT +MBPP.}
	\label{fig:fig12}
\end{figure}

The OSTrack\cite{ostrack}, Stark\cite{stark} (Starks, Starkst), and TransT\cite{transt} trackers are selected to test the effectiveness and generalization of MBPP. We visualized the performance of the four trackers for each attribute in the UOT100 and UTB180 datasets. As shown in Fig. \ref{fig:fig12}, MBPP can effectively improve the performance of different trackers for the two UOT benchmarks. In particular, it can effectively improve performance when it comes to dealing with similar objects, occlusions, low resolution, and fast motion. For the similar object challenges in UOT100, OSTrack, Starks, Starkst, and TransT, respectively, deliver 4.5\%, 6.0\%, 4.3\%, and 1.0\% gains in AUC. These correspond to 2.9\%, 9.7\%, 8.1\%, and 2.0\%, respectively, for the similar object challenges in UTB180. In general, the MBPP is an effective and general method to improve the performance of trackers in challenging similar objects.

Similar objects and low resolution usually cause significant interference, which is appropriately addressed by MBPP. In addition, trajectory prediction can improve tracking performance when a target is temporarily occluded. Since the SOT tracker uses a local search area in fast motion, once tracking fails, the target quickly moves out of the search area and all subsequent tracking fails. In this case, the benefit of correcting the tracking offset in MBPP is significant.

\subsection{Comparison between MBPP and other improved methods}

MBPP is compared with two other methods for improving the tracker (NeighborTrack (represented by OSTrack+N) and Alpha-Refine (represented by OSTrack+AR). In particular, these are compared with OSTrack and the performance of the improved tracker on UOT100 and UTB180 is evaluated. The results are shown in \ref{tab:table4} and Table \ref{tab:table5}.

\begin{table}[h]\scriptsize
	\caption{Comparisons of MBPP with NeighborTrack and Alpha-Refine on the UOT100 dataset}
	\centering
	\begin{tabular}{cccccccc}
  	\cmidrule(r){1-8}	
        \multicolumn{2}{c}{\ }  & \multicolumn{3}{c}{UOT100} &\multicolumn{3}{c}{UOT100’s Similary Subset}\\
 	\cmidrule(r){1-8}
        Tracker&$\bigtriangleup$t/ms & AUC & P & P-Norm & AUC & P & P-Norm\\
 	\cmidrule(r){1-8}
        OSTrack&\ & 66.88 & 62.11 & 84.55 & 58.81 & 54.33 & 72.08\\
        OSTrack+MBPP & +8.04 & 68.40 (+1.52\%) & 63.93 (+1.82\%) & 86.48 (+1.52\%) & 63.27 (+4.46\%) & 59.57 (+5.24\%) & 77.65 (+5.57\%) \\
        \cmidrule(r){1-8}
        OSTrack+N &+21.14 & 67.32 (+0.44\%) & 62.56 (+0.45\%) & 85.23 (+0.68\%) & 59.82 (+1.01\%) & 55.25 (+0.92\%) & 73.63 (+1.55\%)  \\
        OSTrack+N+MBPP &+31.03 & 68.52 (+0.64\%) & 64.00 (+1.89\%) & 86.75 (+2.20\%) & 63.65 (+4.84\%) & 59.86 (+5.53\%) & 78.85 (+6.77\%)  \\
  	\cmidrule(r){1-8}
        OSTrack+AR &+4.79 & 67.32 (+0.44\%) & 60.87 (-1.24\%) & 83.90 (-0.65\%) & 58.53 (-0.28\%) & 53.49 (-0.84\%) & 70.29 (+0.21\%)  \\
        OSTrack+AR+MBPP &+21.03 & 68.36 (+1.48\%) & 61.93 (+0.18\%) & 85.08 (+0.53\%) & 62.05 (+3.24\%) & 57.18 (+2.85\%) & 74.59 (+2.51\%)  \\
  	\cmidrule(r){1-8}
	\end{tabular}
	\label{tab:table4}
\end{table}

\begin{table}[h]\scriptsize
	\caption{Comparisons of MBPP with NeighborTrack and Alpha-Refine on the UTB180}
	\centering
	\begin{tabular}{cccccccc}
  	\cmidrule(r){1-8}	
        \multicolumn{2}{c}{\ }  & \multicolumn{3}{c}{UOT100} &\multicolumn{3}{c}{UOT100’s Similary Subset}\\
 	\cmidrule(r){1-8}
        Tracker&$\bigtriangleup$t/ms & AUC & P & P-Norm & AUC & P & P-Norm\\
 	\cmidrule(r){1-8}
        OSTrack&\ & 63.03 & 56.52 & 72.61  & 54.78  & 50.29 & 62.50\\
        OSTrack+MBPP & +8.04 & 64.13 (+1.10\%) & 56.99 (+0.47\%) & 73.82 (+1.21\%) & 57.70 (+2.92\%) & 52.31 (+2.02\%) & 65.80 (+3.30\%) \\
        \cmidrule(r){1-8}
        OSTrack+N &+21.14 & 64.53 (+1.50\%) & 58.10 (+1.58\%) & 74.16 (+1.55\%) & 58.21 (+3.43\%) & 53.85 (+3.56\%) & 66.29 (+3.79\%)  \\
        OSTrack+N+MBPP &+31.03 & 65.33 (+2.30\%) & 58.29 (+1.77\%) & 74.89 (+2.28\%) & 59.61 (+4.83\%) & 54.33 (+4.04\%) & 67.66 (+5.16\%)  \\
  	\cmidrule(r){1-8}
        OSTrack+AR &+4.79 & 64.88 (+1.85\%) & 59.60 (+3.08\%) & 74.73 (+2.12\%) & 58.74 (+3.96\%) & 55.05 (+4.76\%) & 66.83 (+4.33\%)  \\
        OSTrack+AR+MBPP &+21.03 & 65.55 (+2.52\%) & 59.42 (+2.90\%) & 75.32 (+2.71\%) & 59.96 (+5.18\%) & 55.07 (+4.78\%) & 68.07 (+5.57\%))  \\
  	\cmidrule(r){1-8}
	\end{tabular}
	\label{tab:table5}
\end{table}

The average time required for MBPP is 8 ms, half that of NeighborTrack. MBPP, together with NeighborTrack and Alpha-Refine, can improve tracker performance.

\subsection{Comparison between MBPP and DBPP}

Marine organisms of the same species have a very extremely similar appearance. It is simple for similar objects to resemble the template more closely than the true target when illumination or deformation changes. In this case, similar objects may score higher than the tracked target in the central region of the response map.

\begin{figure}
	\centering
        \includegraphics[width=14cm]{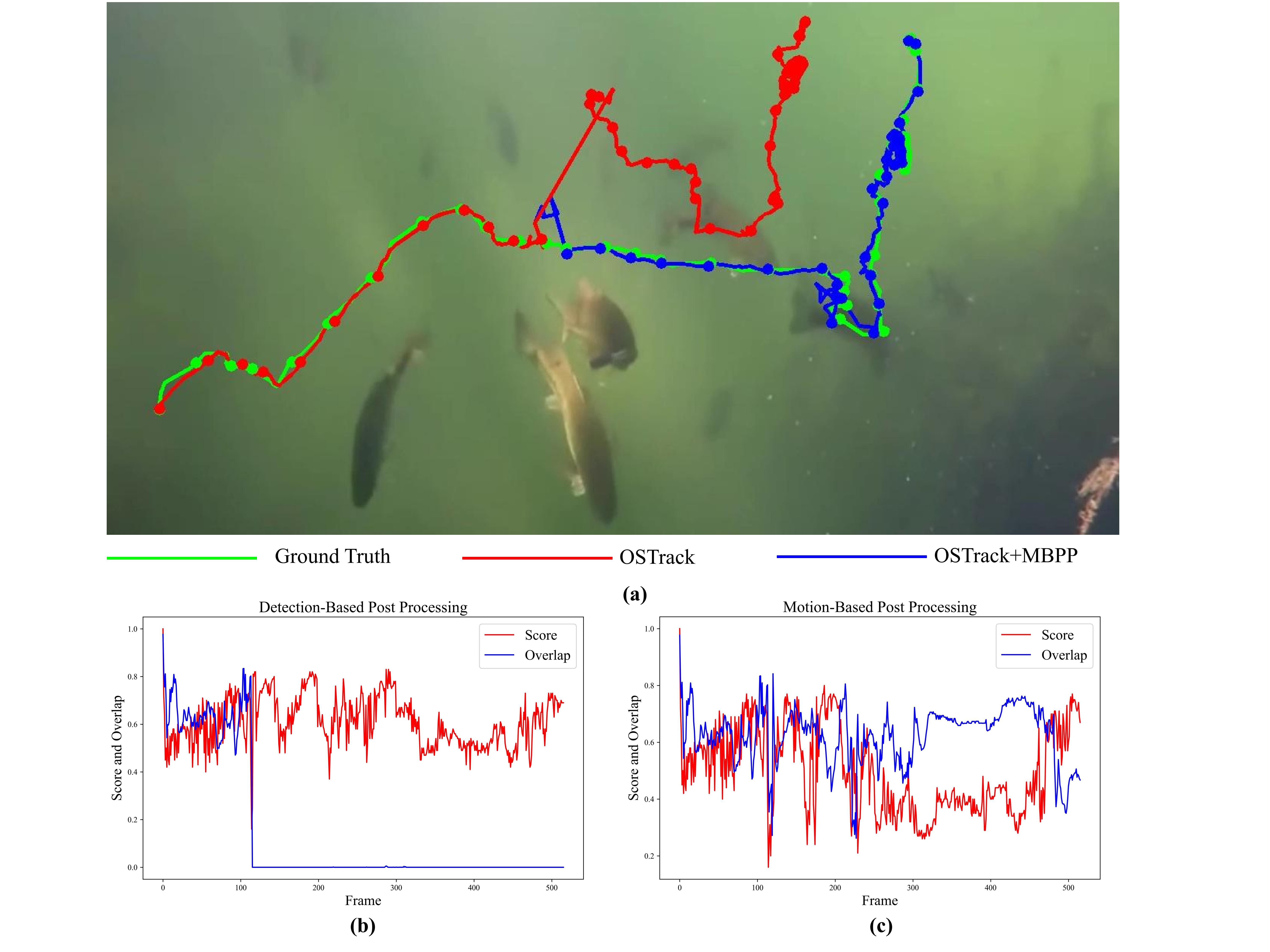}
	\caption{The results of tracking fish in a video sequence in a similar object challenge. (a) The track of the tracking results. (b) The response score and corresponding overlaps of OSTrack. (c) The response score and corresponding overlaps of OTrack+MBPP. The overlap is the IoU between the ground truth and the tracking results.}
	\label{fig:fig13}
\end{figure}

The existing DBPP paradigm is a discrete process that rigidly outputs only the maximum response box in each frame. Although significant errors occur in the process, the detection-based paradigm has no self-correction function. The proposed MBPP paradigm uses not only the maximum score of the response map for the target location, but also the motion information and candidate boxes in each frame. Even if the tracker meets the tracking drift requirements, it can still be effectively corrected. The degree of information utilization is shown in table \ref{tab:table6}, while the effect of MBPP on trajectory is shown in Fig. \ref{fig:fig13}.

\begin{table}
	\caption{The use of information in different post-processing paradigms}
	\centering
	\begin{tabular}{ccccc}

		\cmidrule(r){1-5}
		\multicolumn{1}{c}{\ } & \multicolumn{2}{c}{Single Frame Information} &\multicolumn{2}{c}{Image Sequence Information} \\
  	\cmidrule(r){1-5}
		paradigm & Max response value & Candidate information  & Target motion information & Other information    \\
  	\cmidrule(r){1-5}
		Detection-based       & \checkmark & \  & \  & \      \\
		Motion-based (ours)   & \checkmark & \checkmark  & \checkmark  & \  \\
        \cmidrule(r){1-5}
	\end{tabular}
	\label{tab:table6}
\end{table}

\subsection{Ablation Experiment}

We designed ablation experiments to explore the contribution of each component to UOSTrack. Analysis results included AUC, P, and P-NORM for UOT100 and UTB180. 

\begin{table}[h]\scriptsize
	\caption{Ablation analysis of UHOT and MDPP on UOT100 and UTB180}
	\centering
	\begin{tabular}{cccccccccccc}
  	\cmidrule(r){1-12}	
        \multicolumn{1}{c}{\ }&\multicolumn{2}{c}{Original}&\multicolumn{2}{c}{UOHT}&\multicolumn{1}{c}{MBPP}&\multicolumn{3}{c}{UOT100}&\multicolumn{3}{c}{UTB180} \\
 	\cmidrule(r){1-12}
        Tracker& Other Datasets & COCO &FishExtend &RUOD & AUC & Used & P & P-Norm & AUC & P & P-Norm\\
 	\cmidrule(r){1-12}
        \ &\checkmark &\checkmark &\ &\ &\ & 66.88 & 62.11 & 84.55  & 63.03  & 56.52 & 72.61\\
  	\cmidrule(r){2-12}
        \ &\checkmark &\checkmark &\checkmark &\ &\ & 67.42 & 63.05 & 85.09  & 65.22  & 59.50 & 75.63\\
        \ &\checkmark &\checkmark &\ &\checkmark &\ & 68.05 & 63.38 & 85.95  & 65.67  & 60.01 & 76.44\\
        UOSTrack &\checkmark &\ &\checkmark &\checkmark &\ & 67.29 & 63.68 & 86.13  & 65.15  & 59.29 & 76.01\\
  	\cmidrule(r){2-12}
        \ &\checkmark &\checkmark &\checkmark &\checkmark &\ & 67.80 & 62.61 & 85.17  & 65.84  & 59.64 & 76.17\\
        \ &\checkmark &\checkmark &\ &\ &\checkmark & 68.40 & 63.93 & 86.48  & 64.13  & 56.99 & 73.82\\
        \ &\checkmark &\checkmark &\checkmark &\checkmark &\checkmark & 68.24 & 63.15 & 85.62  & 66.71  & 60.25 & 77.02\\
 	\cmidrule(r){1-12}
	\end{tabular}
	\label{tab:table7}
\end{table}

As shown in table \ref{tab:table7}, OSTrack is the baseline algorithm. The different contributions of the COCO, FishExtend, and RUOD datasets are explored. The use of each dataset improved the performance of OSTrack. Therefore, it is therefore recommended to keep the COCO dataset be retained when using UOHT. However, an exploration of the different contributions of the UOHT and MBPP showed that each component plays an important role in UOSTrack.

\subsection{Marine field experiments}
Marine Field Experiments were conducted in the Dalian sea area, Liaoning Province, China. The range of seawater transparency in the experiments is between 3 m-8 m. The turbidity of the water ranges from 14 NTU to 32 NTU. The Submarine topography includes sand and rocky seabed. The underwater camera is used to capture the image stream, then the image stream is transferred to the processing device, and the image frames are processed in the embedded device NVIDIA Jetson AGX Xavier. The operational height from seabed of the UV is 0.6 m, and the cruising speed is 1.2 knots. The UOSTrack is deployed on the embedded processing device and the UV has started the accuracy and stability test for underwater object tracking. The template size is set to 128×128 and the search area size is set to 256×256. The statistical results show that our proposed UOSTrack could run at a rapid rate of 110 FPS.

\begin{figure}
	\centering
        \includegraphics[width=14cm]{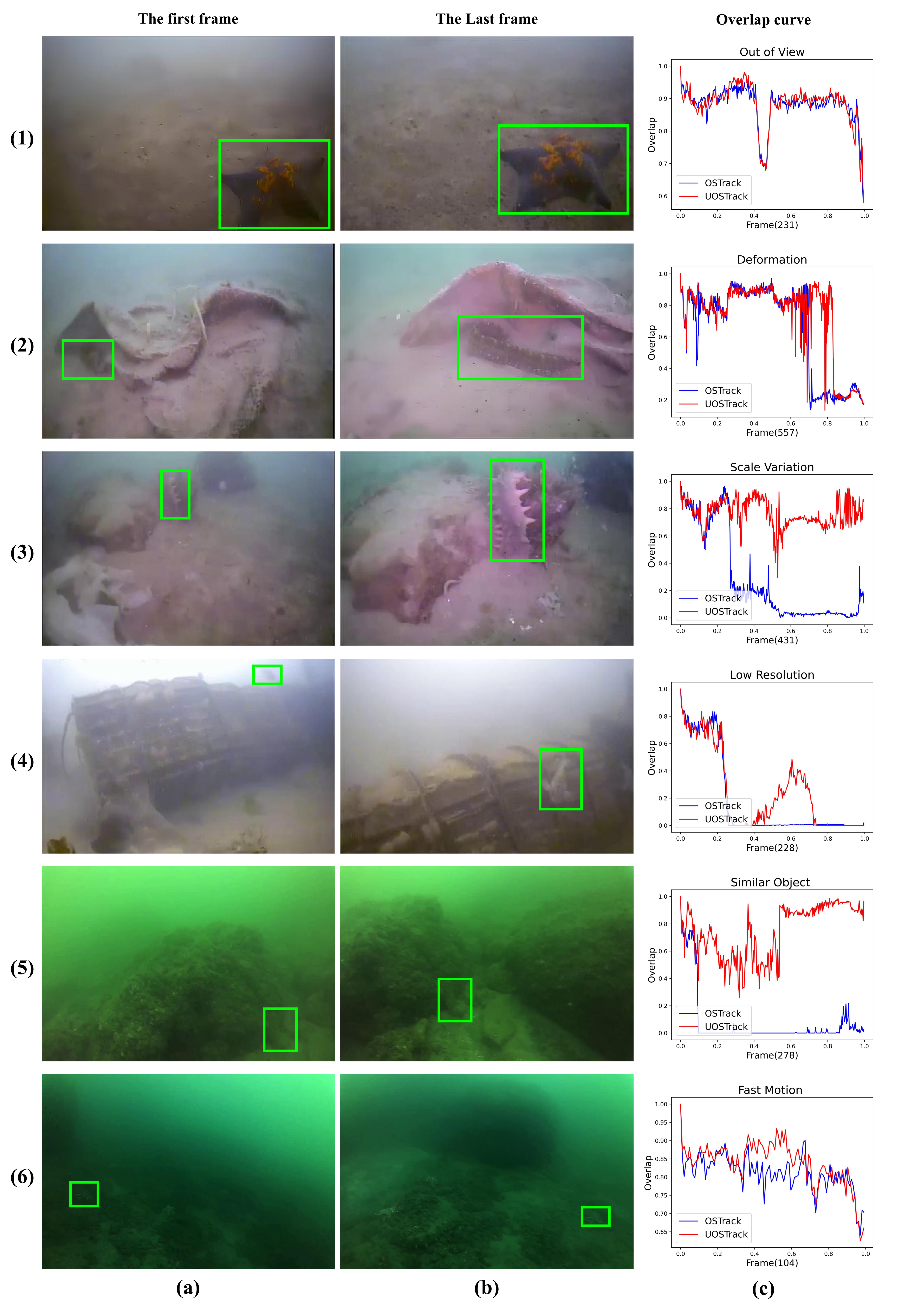}
	\caption{The results of tracking fish in a video sequence in a similar object challenge. (a) The track of the tracking results. (b) The response score and corresponding overlaps of OSTrack. (c) The response score and corresponding overlaps of OTrack+MBPP. The overlap is the IoU between the ground truth and the tracking results.}
	\label{fig:fig14}
\end{figure}

We conducted experiments in 30 typical underwater tracking scenarios. The targets include starfish (7 sequences), sea snakes (2 sequences), sea cucumbers (8 sequences), fishes (8 sequences), and scallops (5 sequences). The experiments include a total of 6 UOT challenges are included in the experiments, each containing a total of 5 sequences. The length of each sequence ranges from 100 to 700 frames. The total frames are 10291. There are 18 sequences reflecting reflects bright underwater images and 12 sequences reflecting reflects green degraded underwater images.

To assess the performance of UOTrack in a real marine environment, some typical underwater tracking challenge sequences are visualized. As shown in Fig. \ref{fig:fig14}, UOSTrack achieves significant improvements over the original OSTrack in the scale variation (Fig. \ref{fig:fig14} (3)) and similar object challenges (Fig. \ref{fig:fig14} (5)), while the UOSTrack slightly outperforms to OSTrack in out-of-view (Fig. \ref{fig:fig14} (1)) and fast motion challenges (Fig. \ref{fig:fig14} (6)). However, in the deformation (Fig. \ref{fig:fig14} (2)) and low-resolution challenges (Fig. \ref{fig:fig14} (4)), the proposed UOSTrack shows unsatisfactory performance but better robustness.

\begin{table}[]
\caption{Comparisons of UOSTrack with OSTrack in the real marine environment}
\centering
\begin{tabular}{c|ccc}
\hline
         & AUC            & P              & P-Norm          \\ \hline
OSTrack  & 42.26          & 47.42          & 47.69           \\ \hline
UOSTrack & 50.21(+7.95\%) & 55.40(+7.98\%) & 58.70(+11.01\%) \\ \hline
\end{tabular}
\label{tab:table8}
\end{table}

\begin{figure}[htbp]
    \begin{minipage}[t]{0.5\linewidth}
        \centering
        \includegraphics[width=7cm]{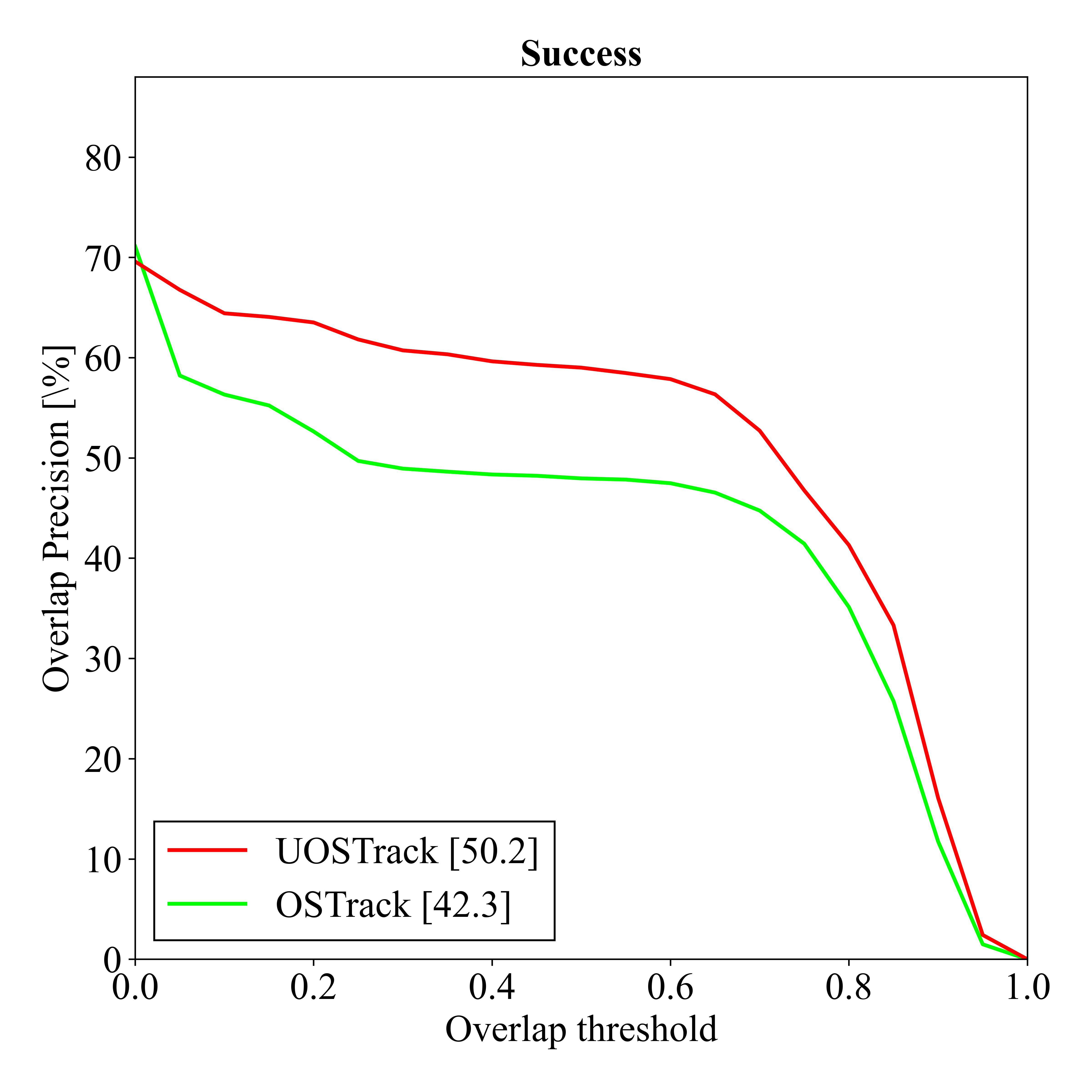}
    \end{minipage}%
    \begin{minipage}[t]{0.5\linewidth}
        \centering
        \includegraphics[width=7cm]{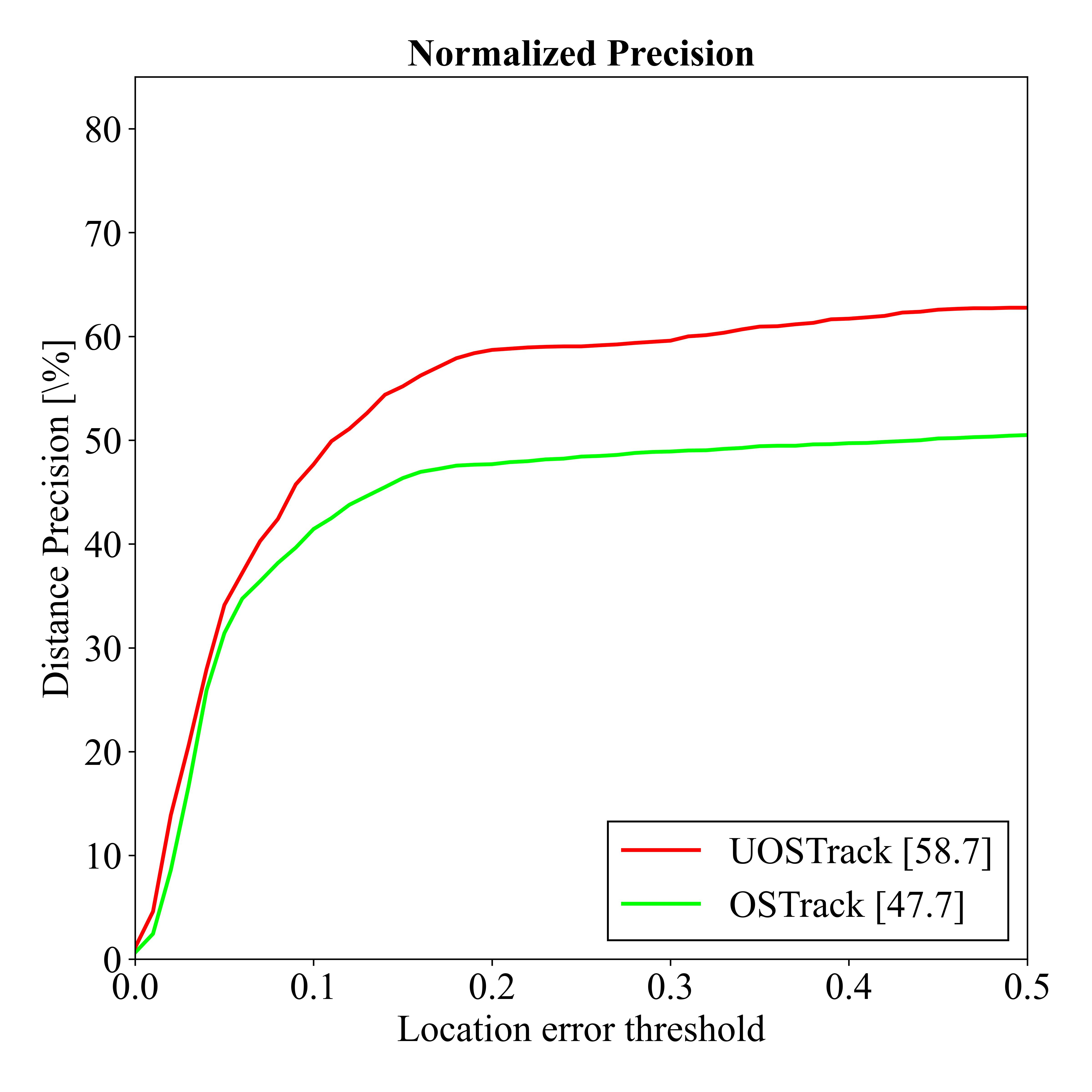}
    \end{minipage}
    \caption{ (a) The curve of the success rate of the test set for the real marine environment. (b) The curve of normalized precision of the real marine environment test set.}
    \label{fig:fig15}
\end{figure}

UOSTrack and OSTrack were evaluated using 30 annotated sequences from real marine environment. As shown in Table \ref{tab:table8}, UOSTrack surpassed OSTrack by 7.95\%, 7.98\%, and 11.01\% in terms of AUC, precision and normalized precision, respectively. The success rate curve and normalized precision curve of UOSTrack and OSTrack are shown in Fig. \ref{fig:fig15}. All significant improvements prove that the proposed UOSTrack achieves much better accuracy and stability for UV operations in the real marine environment. However, the success rate of 50.2\% and the normalized accuracy of 58.7\% indicate that UOSTrack also needs further research to overcome the various challenges of the real marine environment. 

\subsection{Marine organism grasping experiments}
\begin{figure}
	\centering
        \includegraphics[width=14cm]{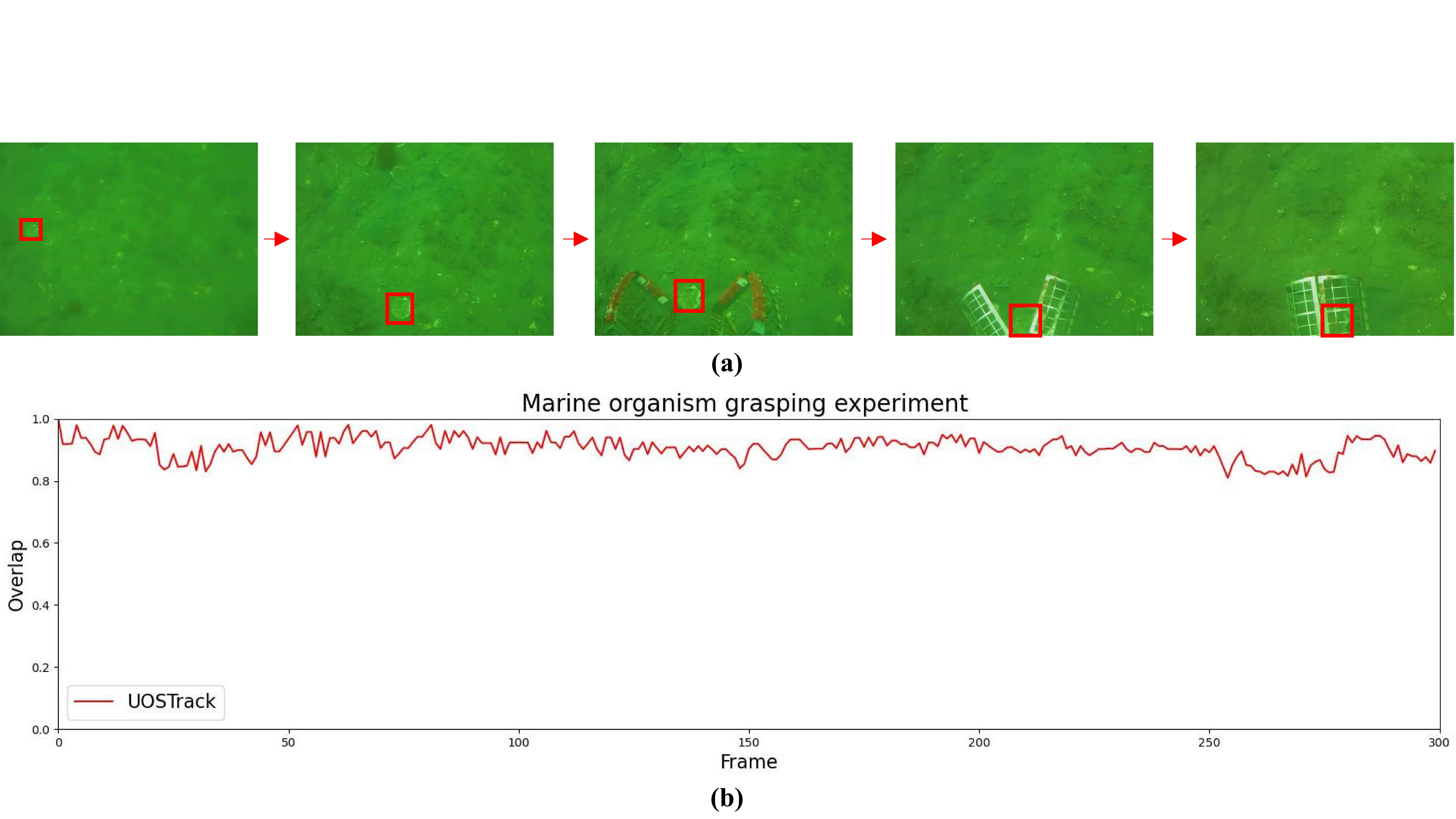}
	\caption{(a) Examples of different frames in grabbing scallop. The red boxes denote the ground truth of the tracked object. (b) The tracking procedure of UOSTrack when grasping the organism by the manipulator on the UV.}
	\label{fig:fig16}
\end{figure}

The performance of the UOSTrack is evaluated in typical grasping tasks for marine organisms. The UOSTrack is used to locate the object at each time point and the object locations in the tracking results are transmitted to the control subsystem, which drives the UV to complete the grasping action with its manipulator. After the manipulator successfully grasps the organism object, the tracker is released for a single test. As shown in Fig. \ref{fig:fig16}, the UOSTrack is able to track the specified object stably and precisely with a minimum overlap of 0.8 during the tracking process.

\section{Conclusion}
In this paper, a real-time UOT tracker named UOSTrack is proposed for marine organism grasping of UVs. UOSTrack utilized the UOHT and MBPP paradigms to achieve cutting-edge performance on two UOT dataset benchmarks and offered adequate tracking performance for deployment on UVs. Extensive experiments have shown the accuracy and stability of UOSTrack. The main contributions are as follows:

(1) The UOHT and MBPP paradigms were designed to address the sample imbalance, improve feature expression, and relocate lost targets in order to obtain greater performance for underwater target tracking. UOSTrack achieves an average performance improvement of 4.41\% maximum when compared with state-of-the-art techniques on two open UOT benchmarks.

(2) Sample imbalance in underwater object tracking has been ignored in the existing research. We designed underwater images and open-air sequence hybrid training (UOHT) paradigm that uniformly adjusts the sample imbalance between the general target set and the underwater target set of underwater trackers during training, which provides a new route for improving underwater object trackers, and the reasons for the effectiveness of UOHT on different attributes is analyzed.

(3) To address the problem of similar objects arising from the aggregation of marine organisms, we designed a simple but effective motion-based post-processing (MBPP) paradigm for predicting trajectories and reusing response map to rectify target tracking drift during the interference caused by similar objects. MBPP is a general and independent method that optimizes the tracking trajectory in similar object challenge.

(4) We have deployed UOSTrack on our developed UV and estimated its performance in the real marine organism grasping tasks of the UV. Marine field experiments verify that UOSTrack has significant performance improvements compared with OSTrack. The improvement in accuracy of UOSTrack is nearly 8\%. The UOSTrack can guide the UV accurately and stably to grasp marine organism objects.

\section{Acknowledgments}
This research was funded by the Natural Science Foundation of Hainan Province, grant number 2021JJLH0002, by the Foundation of National Key Laboratory, grant number JCKYS2022SXJQR-06 and 2021JCJQ-SYSJJ-LB06912, by the National Natural Science Foundation of China, grant number 52071099 and 52071104.

\section{Appendix A}

More details can be found in Table \ref{tab:table9}.

\begin{table}
	\caption{Sequence names of UOT100's similar subsets}
	\centering
	\begin{tabular}{l}
		\toprule
		\multicolumn{1}{c}{UOT100's similarity subsets }                                        \\
		\midrule
                ArmyDiver1, ArmyDiver2, ArmyDiver3, ClickerAndTarget, ColourChangingSquid, CrabTrap,\\
                CressiGuillaumeNeri1, CressiGuillaumeNeri2, Diving360Degree2, FightingEels2, GarryFish,\\
                GiantCuttlefish2, GreenMoreyEel1, GreenMoreyEel3, JerkbaitBites, MantaRescue1, MantaRescue2,\\
                MantaRescue3, MantaRescue4, MississippiFish, MonsterCreature2, MythBusters, PinkFish, \\
                Rocketman, ScubaDiving1, ScubaDiving2, SharkSuckers2, WhiteShark\\
		\bottomrule
	\end{tabular}
	\label{tab:table9}
\end{table}
\bibliographystyle{unsrt}
\bibliography{references}  

\begin{thebibliography}{10}

\bibitem{kcf_based_uot2}
Chaoyu Sun, Zhaoliang Wan, Hai Huang, Guocheng Zhang, Xuan Bao, Jiyong Li,
  Mingwei Sheng, and Xu~Yang.
\newblock Intelligent target visual tracking and control strategy for open
  frame underwater vehicles.
\newblock {\em Robotica}, 39(10):1791--1805, 2021.

\bibitem{wu2023hybrid}
Xiaofeng Wu, Xinyue Han, Zongyu Zhang, Han Wu, Xu~Yang, and Hai Huang.
\newblock A hybrid excitation model based lightweight siamese network for
  underwater vehicle object tracking missions.
\newblock {\em Journal of Marine Science and Engineering}, 11(6):1127, 2023.

\bibitem{kcf_based_uot1}
Xiaojing Li, Zhiqiang Wei, Lei Huang, Jie Nie, Wenfeng Zhang, and Lu~Wang.
\newblock Real-time underwater fish tracking based on adaptive multi-appearance
  model.
\newblock In {\em 2018 25th IEEE international conference on image processing
  (ICIP)}, pages 2710--2714. IEEE, 2018.

\bibitem{bacf_based_uot1}
Ying Lu, Huibin Wang, Zhe Chen, and Zhen Zhang.
\newblock Multi-scale underwater object tracking by adaptive feature fusion.
\newblock In {\em International Symposium on Artificial Intelligence and
  Robotics 2021}, volume 11884, pages 346--357. SPIE, 2021.

\bibitem{uot100}
Karen Panetta, Landry Kezebou, Victor Oludare, and Sos Agaian.
\newblock Comprehensive underwater object tracking benchmark dataset and
  underwater image enhancement with gan.
\newblock {\em IEEE Journal of Oceanic Engineering}, 47(1):59--75, 2021.

\bibitem{ustark}
Yunfeng Li, Wei Huo, Zhuoyan Liu, Bo~Wang, and Ye~Li.
\newblock Ustark: underwater image domain-adaptive tracker based on stark.
\newblock {\em Journal of Electronic Imaging}, 31(5):053012, 2022.

\bibitem{mosse}
David~S Bolme, J~Ross Beveridge, Bruce~A Draper, and Yui~Man Lui.
\newblock Visual object tracking using adaptive correlation filters.
\newblock In {\em 2010 IEEE computer society conference on computer vision and
  pattern recognition}, pages 2544--2550. IEEE, 2010.

\bibitem{keeptrack}
Christoph Mayer, Martin Danelljan, Danda~Pani Paudel, and Luc Van~Gool.
\newblock Learning target candidate association to keep track of what not to
  track.
\newblock In {\em Proceedings of the IEEE/CVF International Conference on
  Computer Vision}, pages 13444--13454, 2021.

\bibitem{ostrack}
Botao Ye, Hong Chang, Bingpeng Ma, Shiguang Shan, and Xilin Chen.
\newblock Joint feature learning and relation modeling for tracking: A
  one-stream framework.
\newblock In {\em European Conference on Computer Vision}, pages 341--357.
  Springer, 2022.

\bibitem{utb180}
Basit Alawode, Yuhang Guo, Mehnaz Ummar, Naoufel Werghi, Jorge Dias, Ajmal
  Mian, and Sajid Javed.
\newblock Utb180: A high-quality benchmark for underwater tracking.
\newblock In {\em Proceedings of the Asian Conference on Computer Vision},
  pages 3326--3342, 2022.

\bibitem{vital}
Yibing Song, Chao Ma, Xiaohe Wu, Lijun Gong, Linchao Bao, Wangmeng Zuo, Chunhua
  Shen, Rynson~WH Lau, and Ming-Hsuan Yang.
\newblock Vital: Visual tracking via adversarial learning.
\newblock In {\em Proceedings of the IEEE conference on computer vision and
  pattern recognition}, pages 8990--8999, 2018.

\bibitem{lu2020deeploss}
Xiankai Lu, Chao Ma, Jianbing Shen, Xiaokang Yang, Ian Reid, and Ming-Hsuan
  Yang.
\newblock Deep object tracking with shrinkage loss.
\newblock {\em IEEE transactions on pattern analysis and machine intelligence},
  44(5):2386--2401, 2020.

\bibitem{dasiamrpn}
Zheng Zhu, Qiang Wang, Bo~Li, Wei Wu, Junjie Yan, and Weiming Hu.
\newblock Distractor-aware siamese networks for visual object tracking.
\newblock In {\em Proceedings of the European conference on computer vision
  (ECCV)}, pages 101--117, 2018.

\bibitem{imagenetdetection}
Olga Russakovsky, Jia Deng, Hao Su, Jonathan Krause, Sanjeev Satheesh, Sean Ma,
  Zhiheng Huang, Andrej Karpathy, Aditya Khosla, Michael Bernstein, et~al.
\newblock Imagenet large scale visual recognition challenge.
\newblock {\em International journal of computer vision}, 115:211--252, 2015.

\bibitem{COCO}
Tsung-Yi Lin, Michael Maire, Serge Belongie, James Hays, Pietro Perona, Deva
  Ramanan, Piotr Doll{\'a}r, and C~Lawrence Zitnick.
\newblock Microsoft coco: Common objects in context.
\newblock In {\em Computer Vision--ECCV 2014: 13th European Conference, Zurich,
  Switzerland, September 6-12, 2014, Proceedings, Part V 13}, pages 740--755.
  Springer, 2014.

\bibitem{IDDSN}
Kun Zhu, Xiaodong Zhang, Guanzhou Chen, Xiaoliang Tan, Puyun Liao, Hongyu Wu,
  Xiujuan Cui, Yinan Zuo, and Zhiyong Lv.
\newblock Single object tracking in satellite videos: Deep siamese network
  incorporating an interframe difference centroid inertia motion model.
\newblock {\em Remote Sensing}, 13(7):1298, 2021.

\bibitem{kcf}
Jo{\~a}o~F Henriques, Rui Caseiro, Pedro Martins, and Jorge Batista.
\newblock High-speed tracking with kernelized correlation filters.
\newblock {\em IEEE transactions on pattern analysis and machine intelligence},
  37(3):583--596, 2014.

\bibitem{dsst}
Martin Danelljan, Gustav H{\"a}ger, Fahad~Shahbaz Khan, and Michael Felsberg.
\newblock Discriminative scale space tracking.
\newblock {\em IEEE transactions on pattern analysis and machine intelligence},
  39(8):1561--1575, 2016.

\bibitem{siamfc}
Luca Bertinetto, Jack Valmadre, Joao~F Henriques, Andrea Vedaldi, and Philip~HS
  Torr.
\newblock Fully-convolutional siamese networks for object tracking.
\newblock In {\em European conference on computer vision}, pages 850--865.
  Springer, 2016.

\bibitem{siamrpn++}
Bo~Li, Wei Wu, Qiang Wang, Fangyi Zhang, Junliang Xing, and Junjie Yan.
\newblock Siamrpn++: Evolution of siamese visual tracking with very deep
  networks.
\newblock In {\em Proceedings of the IEEE/CVF Conference on Computer Vision and
  Pattern Recognition}, pages 4282--4291, 2019.

\bibitem{siamrpn}
Bo~Li, Junjie Yan, Wei Wu, Zheng Zhu, and Xiaolin Hu.
\newblock High performance visual tracking with siamese region proposal
  network.
\newblock In {\em Proceedings of the IEEE conference on computer vision and
  pattern recognition}, pages 8971--8980, 2018.

\bibitem{stark}
Bin Yan, Houwen Peng, Jianlong Fu, Dong Wang, and Huchuan Lu.
\newblock Learning spatio-temporal transformer for visual tracking.
\newblock In {\em Proceedings of the IEEE/CVF International Conference on
  Computer Vision}, pages 10448--10457, 2021.

\bibitem{mixformer}
Yutao Cui, Cheng Jiang, Limin Wang, and Gangshan Wu.
\newblock Mixformer: End-to-end tracking with iterative mixed attention.
\newblock In {\em Proceedings of the IEEE/CVF Conference on Computer Vision and
  Pattern Recognition}, pages 13608--13618, 2022.

\bibitem{aiatrack}
Shenyuan Gao, Chunluan Zhou, Chao Ma, Xinggang Wang, and Junsong Yuan.
\newblock Aiatrack: Attention in attention for transformer visual tracking.
\newblock In {\em European Conference on Computer Vision}, pages 146--164.
  Springer, 2022.

\bibitem{dimp}
Goutam Bhat, Martin Danelljan, Luc~Van Gool, and Radu Timofte.
\newblock Learning discriminative model prediction for tracking.
\newblock In {\em Proceedings of the IEEE/CVF international conference on
  computer vision}, pages 6182--6191, 2019.

\bibitem{tomp}
Christoph Mayer, Martin Danelljan, Goutam Bhat, Matthieu Paul, Danda~Pani
  Paudel, Fisher Yu, and Luc Van~Gool.
\newblock Transforming model prediction for tracking.
\newblock In {\em Proceedings of the IEEE/CVF Conference on Computer Vision and
  Pattern Recognition}, pages 8731--8740, 2022.

\bibitem{tld_based_uot1}
Junhui Wang, Meng Zhao, Li~Zou, Yi~Hu, Xuezhen Cheng, and Xiaofeng Liu.
\newblock Fish tracking based on improved tld algorithm in real-world
  underwater environment.
\newblock {\em Marine Technology Society Journal}, 53(3):80--89, 2019.

\bibitem{pf_based_uot1l}
Deepak~Kumar Rout, Badri~Narayan Subudhi, Thangaraj Veerakumar, and Santanu
  Chaudhury.
\newblock Walsh--hadamard-kernel-based features in particle filter framework
  for underwater object tracking.
\newblock {\em IEEE Transactions on Industrial Informatics}, 16(9):5712--5722,
  2019.

\bibitem{pf_based_uot2}
Pranab~Gajanan Bhat, Badri~Narayan Subudhi, T~Veerakumar, Vijay Laxmi, and
  Manoj~Singh Gaur.
\newblock Multi-feature fusion in particle filter framework for visual
  tracking.
\newblock {\em IEEE Sensors Journal}, 20(5):2405--2415, 2019.

\bibitem{usiamrpn++}
Zongsheng Wang, Jiaqi Wang, and Ruxin Fan.
\newblock An underwater single target tracking method using siamrpn++ based on
  inverted residual bottleneck block.
\newblock {\em IEEE Access}, 9:25148--25157, 2021.

\bibitem{wang2022underwater}
Huibin Wang, Ying Lu, Zhe Chen, Jie Shen, and Min Zhang.
\newblock Underwater object tracking by image enhancement and feature fusion.
\newblock In {\em 2022 International Conference on Computer Engineering and
  Artificial Intelligence (ICCEAI)}, pages 448--450. IEEE, 2022.

\bibitem{wang2018adaptive}
Hesheng Wang, Bohan Yang, Jingchuan Wang, Xinwu Liang, Weidong Chen, and
  Yun-Hui Liu.
\newblock Adaptive visual servoing of contour features.
\newblock {\em IEEE/ASME Transactions on Mechatronics}, 23(2):811--822, 2018.

\bibitem{vit}
Alexey Dosovitskiy, Lucas Beyer, Alexander Kolesnikov, Dirk Weissenborn,
  Xiaohua Zhai, Thomas Unterthiner, Mostafa Dehghani, Matthias Minderer, Georg
  Heigold, Sylvain Gelly, et~al.
\newblock An image is worth 16x16 words: Transformers for image recognition at
  scale.
\newblock {\em arXiv preprint arXiv:2010.11929}, 2020.

\bibitem{lasot}
Heng Fan, Liting Lin, Fan Yang, Peng Chu, Ge~Deng, Sijia Yu, Hexin Bai, Yong
  Xu, Chunyuan Liao, and Haibin Ling.
\newblock Lasot: A high-quality benchmark for large-scale single object
  tracking.
\newblock In {\em Proceedings of the IEEE/CVF conference on computer vision and
  pattern recognition}, pages 5374--5383, 2019.

\bibitem{got10k}
Lianghua Huang, Xin Zhao, and Kaiqi Huang.
\newblock Got-10k: A large high-diversity benchmark for generic object tracking
  in the wild.
\newblock {\em IEEE Transactions on Pattern Analysis and Machine Intelligence},
  43(5):1562--1577, 2019.

\bibitem{trackingnet}
Matthias Muller, Adel Bibi, Silvio Giancola, Salman Alsubaihi, and Bernard
  Ghanem.
\newblock Trackingnet: A large-scale dataset and benchmark for object tracking
  in the wild.
\newblock In {\em Proceedings of the European conference on computer vision
  (ECCV)}, pages 300--317, 2018.

\bibitem{ruod}
Chenping Fu, Risheng Liu, Xin Fan, Puyang Chen, Hao Fu, Wanqi Yuan, Ming Zhu,
  and Zhongxuan Luo.
\newblock Rethinking general underwater object detection: Datasets, challenges,
  and solutions.
\newblock {\em Neurocomputing}, 517:243--256, 2023.

\bibitem{focalloss}
Hei Law and Jia Deng.
\newblock Cornernet: Detecting objects as paired keypoints.
\newblock In {\em Proceedings of the European conference on computer vision
  (ECCV)}, pages 734--750, 2018.

\bibitem{giou}
Hamid Rezatofighi, Nathan Tsoi, JunYoung Gwak, Amir Sadeghian, Ian Reid, and
  Silvio Savarese.
\newblock Generalized intersection over union: A metric and a loss for bounding
  box regression.
\newblock In {\em Proceedings of the IEEE/CVF conference on computer vision and
  pattern recognition}, pages 658--666, 2019.

\bibitem{sort}
Alex Bewley, Zongyuan Ge, Lionel Ott, Fabio Ramos, and Ben Upcroft.
\newblock Simple online and realtime tracking.
\newblock In {\em 2016 IEEE international conference on image processing
  (ICIP)}, pages 3464--3468. IEEE, 2016.

\bibitem{ostrackN}
Yu-Hsi Chen, Chien-Yao Wang, Cheng-Yun Yang, Hung-Shuo Chang, Youn-Long Lin,
  Yung-Yu Chuang, and Hong-Yuan~Mark Liao.
\newblock Neighbortrack: Improving single object tracking by bipartite matching
  with neighbor tracklets.
\newblock {\em arXiv preprint arXiv:2211.06663}, 2022.

\bibitem{transt}
Xin Chen, Bin Yan, Jiawen Zhu, Dong Wang, Xiaoyun Yang, and Huchuan Lu.
\newblock Transformer tracking.
\newblock In {\em Proceedings of the IEEE/CVF Conference on Computer Vision and
  Pattern Recognition}, pages 8126--8135, 2021.

\bibitem{trdimp}
Ning Wang, Wengang Zhou, Jie Wang, and Houqiang Li.
\newblock Transformer meets tracker: Exploiting temporal context for robust
  visual tracking.
\newblock In {\em Proceedings of the IEEE/CVF Conference on Computer Vision and
  Pattern Recognition}, pages 1571--1580, 2021.

\bibitem{siambanacm}
Wencheng Han, Xingping Dong, Fahad~Shahbaz Khan, Ling Shao, and Jianbing Shen.
\newblock Learning to fuse asymmetric feature maps in siamese trackers.
\newblock In {\em Proceedings of the IEEE/CVF Conference on Computer Vision and
  Pattern Recognition}, pages 16570--16580, 2021.

\bibitem{siamban}
Zedu Chen, Bineng Zhong, Guorong Li, Shengping Zhang, and Rongrong Ji.
\newblock Siamese box adaptive network for visual tracking.
\newblock In {\em Proceedings of the IEEE/CVF conference on computer vision and
  pattern recognition}, pages 6668--6677, 2020.

\bibitem{atom}
Martin Danelljan, Goutam Bhat, Fahad~Shahbaz Khan, and Michael Felsberg.
\newblock Atom: Accurate tracking by overlap maximization.
\newblock In {\em Proceedings of the IEEE/CVF Conference on Computer Vision and
  Pattern Recognition}, pages 4660--4669, 2019.

\bibitem{siamcar}
Dongyan Guo, Jun Wang, Ying Cui, Zhenhua Wang, and Shengyong Chen.
\newblock Siamcar: Siamese fully convolutional classification and regression
  for visual tracking.
\newblock In {\em Proceedings of the IEEE/CVF conference on computer vision and
  pattern recognition}, pages 6269--6277, 2020.

\bibitem{tld}
Zdenek Kalal, Krystian Mikolajczyk, and Jiri Matas.
\newblock Tracking-learning-detection.
\newblock {\em IEEE transactions on pattern analysis and machine intelligence},
  34(7):1409--1422, 2011.

\bibitem{eco}
Martin Danelljan, Goutam Bhat, Fahad Shahbaz~Khan, and Michael Felsberg.
\newblock Eco: Efficient convolution operators for tracking.
\newblock In {\em Proceedings of the IEEE conference on computer vision and
  pattern recognition}, pages 6638--6646, 2017.

\bibitem{bacf}
Hamed Kiani~Galoogahi, Ashton Fagg, and Simon Lucey.
\newblock Learning background-aware correlation filters for visual tracking.
\newblock In {\em Proceedings of the IEEE international conference on computer
  vision}, pages 1135--1143, 2017.

\end{thebibliography}






\end{document}